%% file: main.tex
\documentclass{article}
\usepackage{spconf,amsmath,graphicx}

\usepackage{graphics} 
\usepackage[english]{babel}
\usepackage{blindtext}
\usepackage{float}
\usepackage{tabularx}
\usepackage{amsmath}
\usepackage{graphicx}
\usepackage{caption}
\usepackage{subcaption}
\usepackage{soul}
\usepackage{pgfplots}
\usepackage{subcaption}

\newlength\figureheight 
\newlength\figurewidth 

\title{Robust features for facial action recognition}

\name{Nadav Israel$^{\star}$ \qquad Lior Wolf$^{\star}$ \qquad Ran Barzilay$^{\dagger}$ \qquad Gal Shoval$^{\dagger}$}
\address{$^{\star}$ The Blavatnik School of Computer Science, Tel Aviv University, Israel \\$^{\dagger}$ Sackler Faculty of Medicine, Tel Aviv University, Israel
\\$^{\dagger}$Geha Mental Health Center, Petah Tiqva, Israel}

\begin{document}

\maketitle
\thispagestyle{empty}
\pagestyle{empty}


\begin{abstract}
Automatic recognition of facial gestures is becoming increasingly important as real world AI agents become a reality. In this paper, we present an automated system that recognizes facial gestures by capturing local changes and encoding the motion into a histogram of frequencies.
We evaluate the proposed method by demonstrating its effectiveness on spontaneous face action benchmarks: the FEEDTUM dataset, the Pain dataset and the HMDB51 dataset. The results show that, compared to known methods, the new encoding methods significantly improve the recognition accuracy and the robustness of analysis for a variety of applications.
\end{abstract}

\begin{figure*}[h]
\centering
\begin{subfigure}{0.9\textwidth}
\centering 
\setlength\figureheight{2.25cm} 
\setlength\figurewidth{\textwidth} 
\input{face.tikz} 
\caption{~}
\label{fig:align}
\end{subfigure}

\centering
\begin{subfigure}{0.7\linewidth}

\setlength\figureheight{1.1cm} 
\setlength\figurewidth{\textwidth}

\input{time.tikz} 
\caption{~}
\label{fig:time}
\end{subfigure}
\caption{Face preprocessing. (a) The alignment process is performed by localizing three fiducial points inside the detection crop. (b) illustrates the signal $x_r$ captured from the left eye region over 25 frames.}
\label{fig:preprocess}
\centering

\end{figure*}
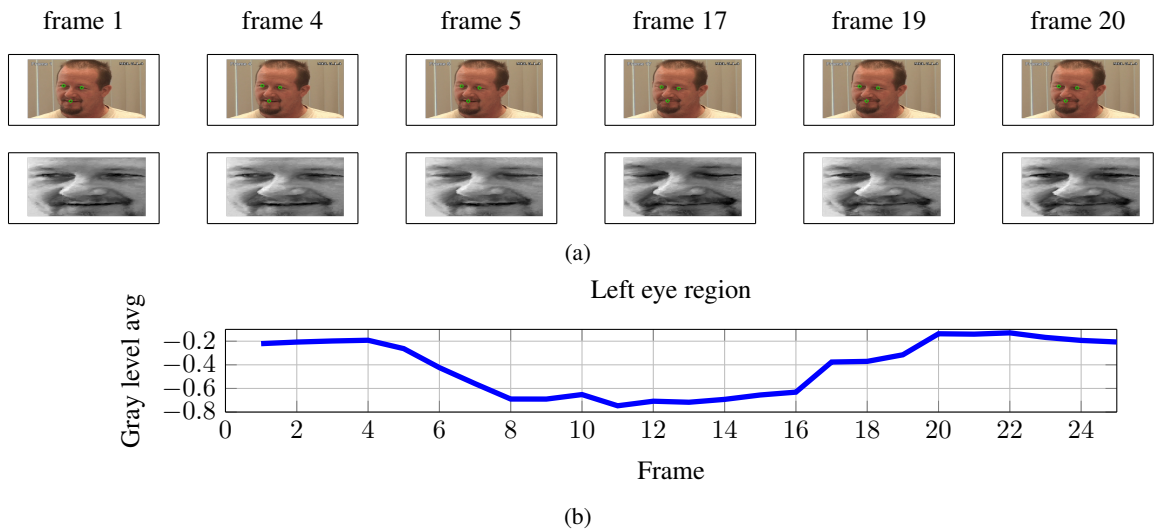

\section{Introduction}
Face action recognition has been an active research domain that has made great strides over the past decade~\cite{c1}. In many real-world applications, the goal is to recognize or infer intention or other psychological states rather than facial actions alone. For this purpose, both narrowing the number of facial actions of interest and paying attention to the video context may be critical to the success of an automated system. A face action recognition system is normally composed of four main steps: (i) face detection and tracking, (ii) face alignment, (iii) feature extraction, and (iv) classification. In our framework, we track face regions using the commonly used Viola–Jones object detection algorithm~\cite{c2}. The second step in the framework is face alignment~\cite{c3}, which is based on locating semantic facial landmarks such as the eyes, nose, mouth and chin. This step is also standard and is an essential part of face recognition, face animation, and 3D face modeling. The third step in the framework is feature extraction, which is the area where we make our contributions. 
Choosing suitable feature extraction and feature selection algorithms play the central roles in providing discriminative and robust information, particularly for spontaneous facial action that are occurs with changes in pose and illumination, and face movements. Accordingly, the features should be extracted in a way that is robust to these external changes. 
The optimal features should minimize within class variations of expressions, while maximizing the signal between class variations. If inadequate features are used, even the best classifier would fail to achieve accurate recognition~\cite{c5}. As we experimentally demonstrate, generic action recognition features, even successful ones, are not suitable for the task at hand. 
\noindent We present a novel method for facial features extraction which is based on capturing local changes and encoding the facial motion into a histogram of frequencies in order to classify the face actions.
To date, there are two common ways to recognize facial expressions. Those include direct recognition of prototypical expressions (as described in this paper) and recognition of expressions through a facial action coding system (FACS) by the Ekman method~\cite{c6,c7,c8}. Our method belongs to the first type. In an extensive set of experiments, we demonstrate the utility of the proposed method of multiple spontaneous facial action datasets: FEEDTUM~\cite{c9}, PainDB~\cite{c17} and HMDB~\cite{c12}. In all three cases, we demonstrate state of the art results. 


\section{Related work}
\looseness=-1
In the recent past, many different approaches for facial expression classification in videos have been evaluated. A significant contribution to on research on spontaneous expressions was the introduction of UNBC-McMaster Shoulder Pain dataset~\cite{c17}, which involves subjects experiencing shoulder pain in a clinical setting. A.~Ashraf et al.~\cite{c16} used an approach starting with extracting Active Appearance Model~\cite{c18} based features from each frame and using these to cluster the frames, thus creating training data with size that is manageable by a SVM. P. Lucey et al.~\cite{c11} extended this work by borrowing ideas from the related field of visual speech recognition and proposed to compress the signal in the spatial rather than temporal domain using the Discrete Cosine Transform (DCT). K.~Sikka et al.~\cite{c1} suggested a method called MS-MIL which is proposed to jointly detect and locate pain events in video. In their framework, each video sequence is represented as a bag of multiple segments, and Multiple Instance Learning (MIL) is employed in order to deal with this weakly labeled data in the form of sequence level ground truth.
F. Wallhoff et al.~\cite{c9} discussed innovative holistic and self organizing approaches for efficient facial expression analysis. Their experiments are based on the  publicly available FEEDTUM dataset and achieved accuracy of 61.67\% by using macro motion blocks and Sequential-Floating-Forward-Search (SFFS) based feature selection with use of the aimed at SVM (SVM-SFFS).  More recent methods achieve considerably higher accuracies ~\cite{c1}.
Another central family, that uses low-level representation schemes of the information in a video is Optical-flow based methods. Current state of the art methods are based on dense trajectories~\cite{c14}. The trajectories are extracted efficiently with optical flow and represent the local motion information in the video. The descriptors are then computed as HOG, HOF or MBH on a spatio-temporal volume defined by the trajectory. This method is state of the art on the HMDB51~\cite{c12} dataset. In our work, we only care about seven facial categories out of the 51 classes: chew, drink, eat, laugh, smile, smoke and talk.


\section{Method}
The quality of the extracted features representing the video content plays a key role for the latter classification task. The proposed system is designed to be not only accurate but also rapid and support real-time analysis applications. 
The proposed representation is based on capturing local changes and encoding these local motions into a histogram of frequencies. In this approach, a face video is modeled as a sequence of histograms by the following procedure: (1) An input face image is cropped and normalized as described in Figure~\ref{fig:preprocess}; (2) Each normalized frame is divided into a grid of equally sized cells; (3) The mean graylevel intensity of each cell $r$ at frame $n$ is recorded as $x_r[n]$; (4) For the entire video, the frequency histogram is computed for each cell by using the wavelet transform.

\subsection{Preprocessing}
For the purpose of face action recognition, we track the facial features robustly and efficiently. The appearance of facial features changes due to pose, lighting, and facial expressions making the task difficult and complex. Automatic face registration can be achieved by face detection~\cite{c2} and facial landmarks detection~\cite{c3}. Facial images were cropped and aligned from original frames based on the two eyes and mouth locations. 
\noindent There are several techniques available for performing face detection. In this paper, we have used Viola and Jones face detection technique based on the AdaBoost algorithm. 
\noindent In order to align the faces and to obtain 2D pose normalization, the faces are transformed such that a fixed distance between the two eyes and mouth is achieved. First, we detect three fiducial points inside the detected face region: the center of each eye and the mouth location as illustrated in Figure~\ref{fig:preprocess}. Then, a similarity transformation (scale, rotation and translation) is used to bring these points to three pre-defined anchor points. The fiducial points generated by Intraface~\cite{c3} are used for detecting the face landmarks. ~\cite{c3} used the cascade regression with SIFT feature, and interpreted the cascade regression procedure from a gradient descent view. We reduce the number of fiducial points to three by averaging the detected landmarks of each face part: left eye, right eye and mouth.

\subsection{Filter banks}
\begin{figure}[H]
\centering
\includegraphics[width=8cm,height=1.25cm]{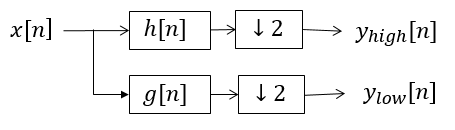}
\caption{A block diagram showing a single level of filter bank analysis.}
\label{fig:filter}
\end{figure}

\begin{figure*}
\centering
  \includegraphics[width=0.8\textwidth,height=2.5cm]{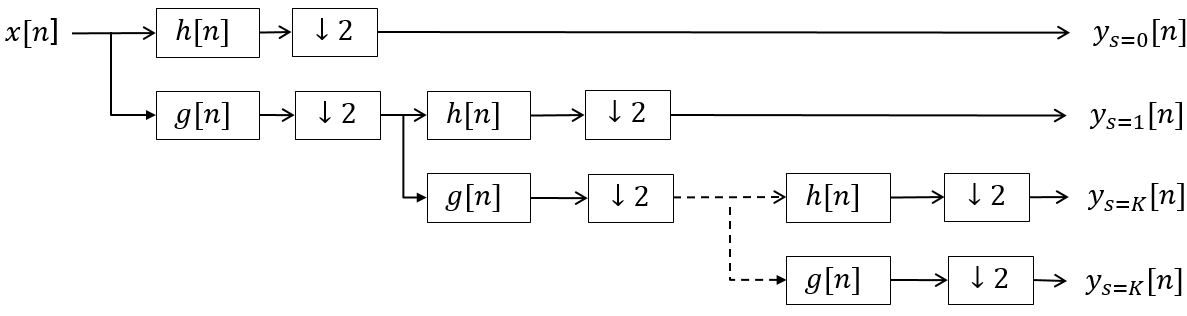}
    \caption{A block diagram of the filter bank depicting a Haar tree structure with K cascade level}
  \label{fig:bank}
\end{figure*}

In the classical applications of multi-rate filter banks, a bank of filters is applied to a discrete input signal and then down-sampled at a fixed rate to produce a set of sub-band signals. The signal is decomposed simultaneously using a high-pass filter h and a low-pass filter g as illustrated in Figure~\ref{fig:filter}. The specific form of filters used is known as quadrature mirror filters~\cite{c19}
This paper employs the uniform K-channel Haar filter banks~\cite{c29} for which the sampling rate is reduced by two in all sub-bands. Figure~\ref{fig:bank} illustrates the filter bank. 
In discrete filter banks, the scale and resolution vary, providing a detailed description of the signal. 
Figure~\ref{fig:bank} depicts the analysis process, in which the signal is analyzed at multiple resolutions. This decomposition has halved the time resolution at each cascade level. However, each output has half the frequency band of the input and the frequency resolution has been doubled. Moreover, computing a complete convolution $x*g$ with subsequent down-sampling would waste computation time. The Lifting scheme is an optimization where these two computations are interleaved. This decomposition is represented as a binary tree with nodes representing a sub-space or sub-scale space with a different time-frequency localization. The tree defines the filter bank used. 

Where $y_{\mathrm{low}}$ and $y_{\mathrm{high}}$ denote as:
\begin{equation*}
y_{\mathrm{low}} [n] = \sum\limits_{k =  - \infty }^\infty  {x[k] g[2n - k]}
,\quad 
y_{\mathrm{high}} [n] = \sum\limits_{k =  - \infty }^\infty  {x[k] h[2n - k]}
\end{equation*}
\begin{equation*}
h[n] = \left[\frac{-\sqrt{2}}{2}, \frac{\sqrt{2}}{2}\right]  
,\quad 
g[n] = \left[\frac{\sqrt{2}}{2}, \frac{\sqrt{2}}{2}\right]
\end{equation*}

\subsection{Histogram of frequencies} \label{sssec:hof}

\begin{figure*}[h]
\centering

\begin{subfigure}{\textwidth}
\centering 
\setlength\figureheight{1.22cm} 
\setlength\figurewidth{0.8\textwidth} 
\input{timeAll.tikz} 
\caption{~}
\label{fig:timeAll}
\end{subfigure}

\begin{subfigure}[b]{\textwidth}    
\centering 
\setlength\figureheight{1.22cm} 
\setlength\figurewidth{0.8\textwidth}
\input{filterBankRes.tikz}       
\caption{~}
\label{fig:filterBanksRes}
\end{subfigure}

\begin{subfigure}[b]{\textwidth}    
\centering 
\setlength\figureheight{1.22cm} 
\setlength\figurewidth{0.8\textwidth}
\input{filterBankHist.tikz}       
\caption{~}
\label{fig:hof}
\end{subfigure}
    
\caption{Illustration of the filter banks processing: (a) The signals $x_r$ recorded from the left eye region for a Pain video (blue) and a No pain  video (red). (b) The matching filter banks coefficients $y_r^s$.  (c) The corresponding histograms $z_r^s=hist(y_r^s)$}.
\label{fig:filterbank}

\end{figure*}
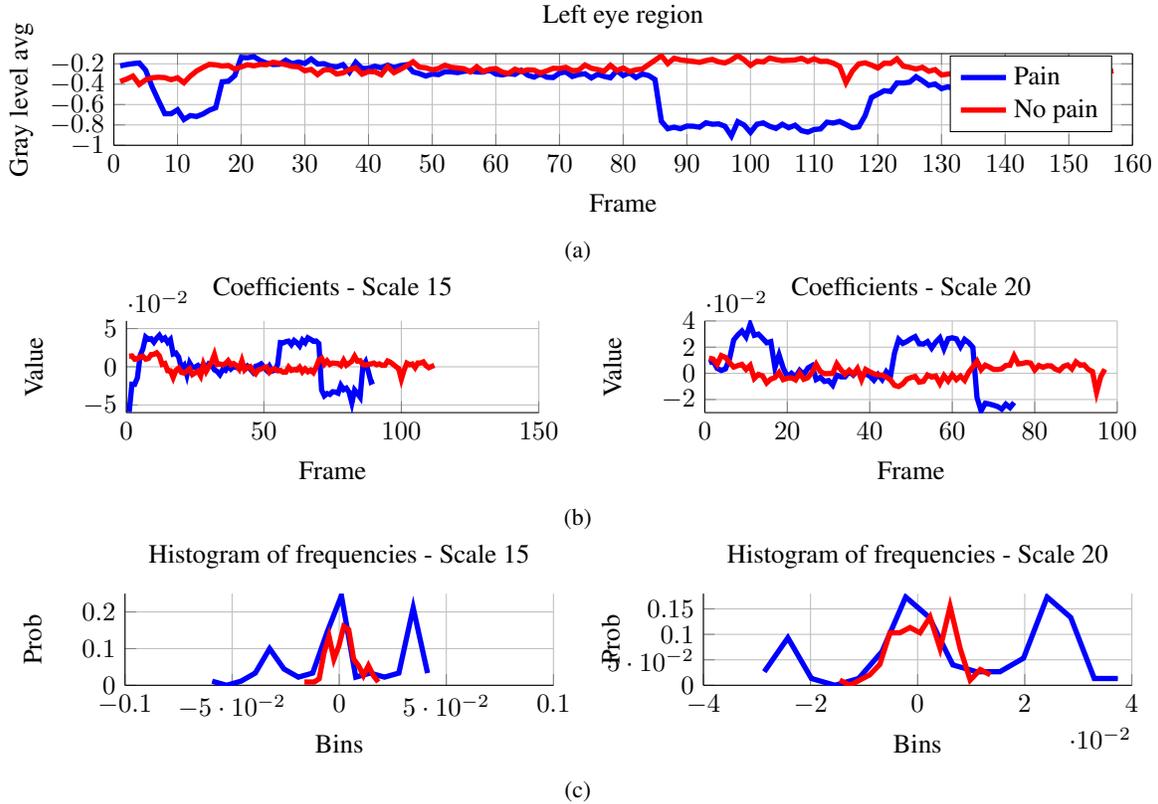

The Haar filter banks are used to decompose a signal into components of different levels of detail. Each such component contains only a certain band of frequencies extracted from the input signal, and the decomposition results in band-pass filtering~\cite{c20, c21}. For example, a usual 2-channel filter bank uses low-pass and high-pass filters in combination with up- and down-samplers to produce the “coarse” and “detailed” components.
In order to obtain a multi-scale video representation that is invariant to shifts in time, we apply the Haar decomposition to multiple time series. Each series is obtained as a vector of average intensities over image regions, $x_r[n]$, where $n$ is the frame index, and $r$ is the regions index. Each region is a cell obtained by fixing a grid over the image.
After the filter bank is applied, we obtain, for each region $r$, coefficients $y_r^s$ for each cascade level $s$ on the Haar tree structure.
In order to represent face gestures regardless of the time at which the action occurs and keep on a compact representation, we collect the coefficients $y_r^s$ of each cascade level into histogram $z_r^s=hist(y_r^s)$. Figure~\ref{fig:filterbank} illustrates the coefficients and histograms in two cascades level.
The combined histogram of frequencies is constructed by concatenating histogram, $z_r^s$, for all cascade levels, $s$, $s=1..K$ for a given region. We define the operator that transforms the signal $x_r$, at region $r$, to a histogram of frequencies as: $p_r = HOF(x_r)$.

The procedure is extremely efficient. Given a face image of size $100 \times 100$, we calculate a convolution with a window size of $5 \times 5$ and  decimate the result to $20 \times 20$ pixels. This takes 0.007 second. Then, the computation in time takes 0.276845  seconds for a video of 256 frames. The time complexity of the Haar filter banks satisfies $T(N) = 2N + T(N/2)$, which leads to an $O(N)$ time for the entire operation, where $N$ is the number of frames in a video.

\section{Classifier}
Bag of words~\cite{c27, c28} models represent a videos as an order-less collection of local features. 
These models have shown remarkable performance in multiple domains including scene recognition~\cite{c22, c23}, object and texture categorization~\cite{c24, c25}, and human activity recognition~\cite{c26}. 
Let $P^j = \left\{ p_1^j, p_2^j,···, p_{N_j}^j \right \}$ denote a set of D-dimensional local features that are extracted from a video $V_j$, where $N_j$ is number of regions in a video $j$.
To encode these features we use the BoW method, with a trained codebook $B \in R^{^{d\times c}}$. The codebook is constructed by quantizing using K-means clustering local feature vectors randomly selected from the videos. 
This encoding method is based on finding, for each local feature, the nearest cluster center. This coding step maps the local feature $p_i^j \in R^d$ to the coefficient $u_i^j \in R^c$. In order to obtain video-level representation, the coding coefficients of all local features in an video are pooled together.
To increase precision, we initialize k-means 8 times and keep the result with the lowest error. Local features are assigned to their closest trained codebook using Euclidean distance. The resulting histograms of word occurrences are used as video descriptors.
\looseness=-1
A linear support vector machine (SVM) was used as the classifier for the proposed method. Specifically, we use the implementation of LIBSVM~\cite{c15}. SVM makes binary decisions, and the multiclass classification here is accomplished by using the one-versus-all rule: a classifier is learned to separate each class from the rest, and a test video is assigned the label of the classifier with the highest response.


\section{Results}
\looseness=-2
We compare the Histogram of frequencies method to the performance of two leading state of the art methods: dense trajectories~\cite{c14} and MIL~\cite{c1}.  In order to evaluate the dense trajectories method, we used the published code~\cite{c14} and employed the default parameters. For most of the datasets, dense trajectories results are significantly lower than the other methods. We have reported results for MIL~\cite{c1} as were reported by authors in using their own implementation. Three public datasets was utilized for our experiments: PainDB, HMDB51 and FEEDTUM. The performance statistics for prescribed experiments are shown in Tables~\ref{table:PainDB},~\ref{table:HMDB51},~\ref{table:FEEDTUM}, respectively.

\looseness=-2
\noindent\textbf{The PainDB dataset}~\cite{c20} included 200 sequences from 25 subjects. Each subject was undergoing some kind of shoulder pain and was asked to perform a series of active and passive movements of their affected and unaffected limbs. Active tests were self-initiated shoulder movements and in passive tests the physiotherapist was responsible for the movement. These sequences were then coded on a number of levels by experts. Following the protocol proposed, labels were binarized into ‘pain’ and ‘no pain’ by defining training instances with OPI $\geq$ 3 (Observer Pain Intensity)) as the positive class (pain) and OPI = 0 as the negative class (no-pain). In the experiments, it is common to include only those subjects which had a minimum of one trial with an OPI rating of 0 (no pain) and one trial with an OPI rating of either 3, 4 or 5 (pain). Intermediate pain intensities of 1 and 2 were omitted, per the protocol in~\cite{c1, c17}. This yielded 147 sequences from 23 subjects for our experiments.
To check for subject generalization, a leave-one-subject out strategy was employed. Thus, there was no overlap of subjects between the training and testing set. 
The classification task focuses on pain / no pain predictions at the video-level. For reporting the results, we followed the strategy employed in~\cite{c1, c16, c17}, where they reported total classification rate or accuracy, which refers to the percentage of correctly classified sequences, computed at the Equal Error Rate (EER) point of the Receiver Operation Curve (ROC).
The comparison among all tested methods is presented in Table~\ref{table:PainDB}. The results show that the proposed method outperforms all literature baselines.

\looseness=-2
\noindent\textbf{The HMDB51 dataset}~\cite{c12} has 51 action classes with a total of 6,766 videos. Each class has more than 100 videos.  All of the videos are obtained from real world scenarios such as movies and YouTube clips. The intra-class variation is very high due to many factors, such as viewpoint, scale, background, illumination etc. In this work, we only consider seven classes (chew, drink, eat, laugh, smile, smoke and talk) that relate to facial action. We also limit the dataset to those videos in which a face was detected in 60\% of the frames. The total number of included videos is 484. All experiments were conducted in leave-one-out fashion. The results are depicted in Table~\ref{table:HMDB51}. The Histogram of frequencies result achieves a significant improvement over the baseline method, which is among the most successful methods on the full HMDB dataset. This supports our notion that face action recognition requires a dedicated set of specialized techniques.

\looseness=-1
\noindent\textbf{The FEEDTUM dataset}~\cite{c9} is a facial expression dataset that consists of 320 videos from 19 subjects showing six basic emotions, namely: anger, disgust, fear, happiness, sadness and surprise. The dataset exhibits natural expressions, which were elicited by showing the subjects several carefully selected video stimulus. Here, similar to the PainDB benchmark, we conduct leave-one-subject out experiments. The results are shown in Table~\ref{table:FEEDTUM}. Our histogram of frequencies method achieves mean classification accuracy of 81.12\% compared to the leading method on this benchmark, which is MS-MIL at 84.50\%.

\begin{table}[h]
\centering
\caption{Comparison of various of methods on the PainDB dataset. The accuracy refers to the percentage of correctly classified sequences, computed at Equal Error Rate (EER) in the Receiver Operation Curve (ROC).}
\label{table:PainDB}
\begin{tabular}{llll}
\hline
Method &  & Accuracy\\
\hline
Dense trajectory [14] &  & 80.70\\
BoW+Max+SVM [13] &  & 81.52\\
MS-MIL [1] &  & 83.70\\
Histogram of frequencies (ours) &  & 85.96\\
\hline
\end{tabular}

\bigskip
\centering
\caption{Comparison of our method to the dense trajectories method on the HMDB51 face subset.}
\label{table:HMDB51}
\begin{tabular}{llll}
\hline
Method &  & Accuracy\\
\hline
Dense trajectory [14] &  & 66.74\\
Histogram of frequencies (ours) &  & 84.71\\
\hline
\end{tabular}

\bigskip
\centering
\caption{Comparison of various methods on the FEEDTUM dataset. The mean classification accuracy over all expressions is reported.}
\label{table:FEEDTUM}
\begin{tabular}{llll}
\hline
Method &  & Accuracy\\
\hline
Dense trajectory [14] &  & 51.38\\
MS-MIL [1] &  & 84.50\\
MilBoost [1] &  & 81.78\\
Histogram of frequencies (ours) &  & 81.12\\
\hline
\end{tabular}
\end{table}


\section{Discussion}
\looseness=-2
This paper outlines a scheme for face action recognition based on a K-level Haar wavelets decomposition. It demonstrates an improvement in recognition accuracy over other popular methods. 
From our experiments, it is evident that spontaneous expressions in videos is a challenging problem due owing to the variability associated with the expression of facial gestures by different subjects at different times and scenarios.
Our results show that a meaningful face action recognition framework can be built based on filtering tools such as the Wavelet transform, which differs considerably from both the existing face action recognition methods and the current action recognition methods.


\cleardoublepage\cleardoublepage

\end{document}

%% file: face.tikz
%
%
\begin{tikzpicture}

\begin{axis}[%
width=0.125\figurewidth,
height=0.419\figureheight,
at={(0\figurewidth,0.581\figureheight)},
scale only axis,
axis on top,
separate axis lines,
every outer x axis line/.append style={black},
every x tick label/.append style={font=\color{black}},
xmin=0.5,
xmax=1200.5,
xtick={\empty},
every outer y axis line/.append style={black},
every y tick label/.append style={font=\color{black}},
y dir=reverse,
ymin=0.5,
ymax=900.5,
ytick={\empty},
axis background/.style={fill=white},
title={frame 1}
]
\addplot [forget plot] graphics [xmin=0.5,xmax=1200.5,ymin=0.5,ymax=900.5] {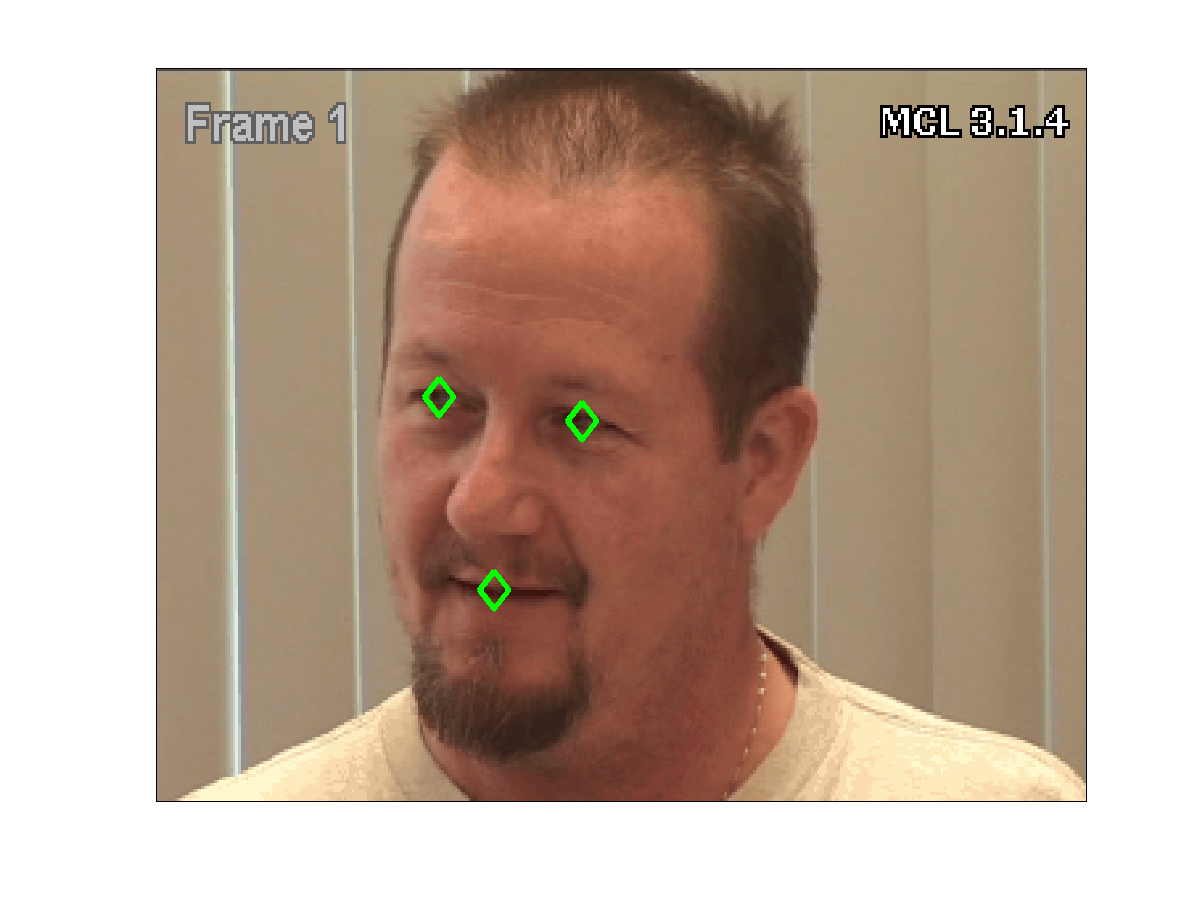};
\end{axis}

\begin{axis}[%
width=0.125\figurewidth,
height=0.419\figureheight,
at={(0.165\figurewidth,0.581\figureheight)},
scale only axis,
axis on top,
separate axis lines,
every outer x axis line/.append style={black},
every x tick label/.append style={font=\color{black}},
xmin=0.5,
xmax=1200.5,
xtick={\empty},
every outer y axis line/.append style={black},
every y tick label/.append style={font=\color{black}},
y dir=reverse,
ymin=0.5,
ymax=900.5,
ytick={\empty},
axis background/.style={fill=white},
title={frame 4}
]
\addplot [forget plot] graphics [xmin=0.5,xmax=1200.5,ymin=0.5,ymax=900.5] {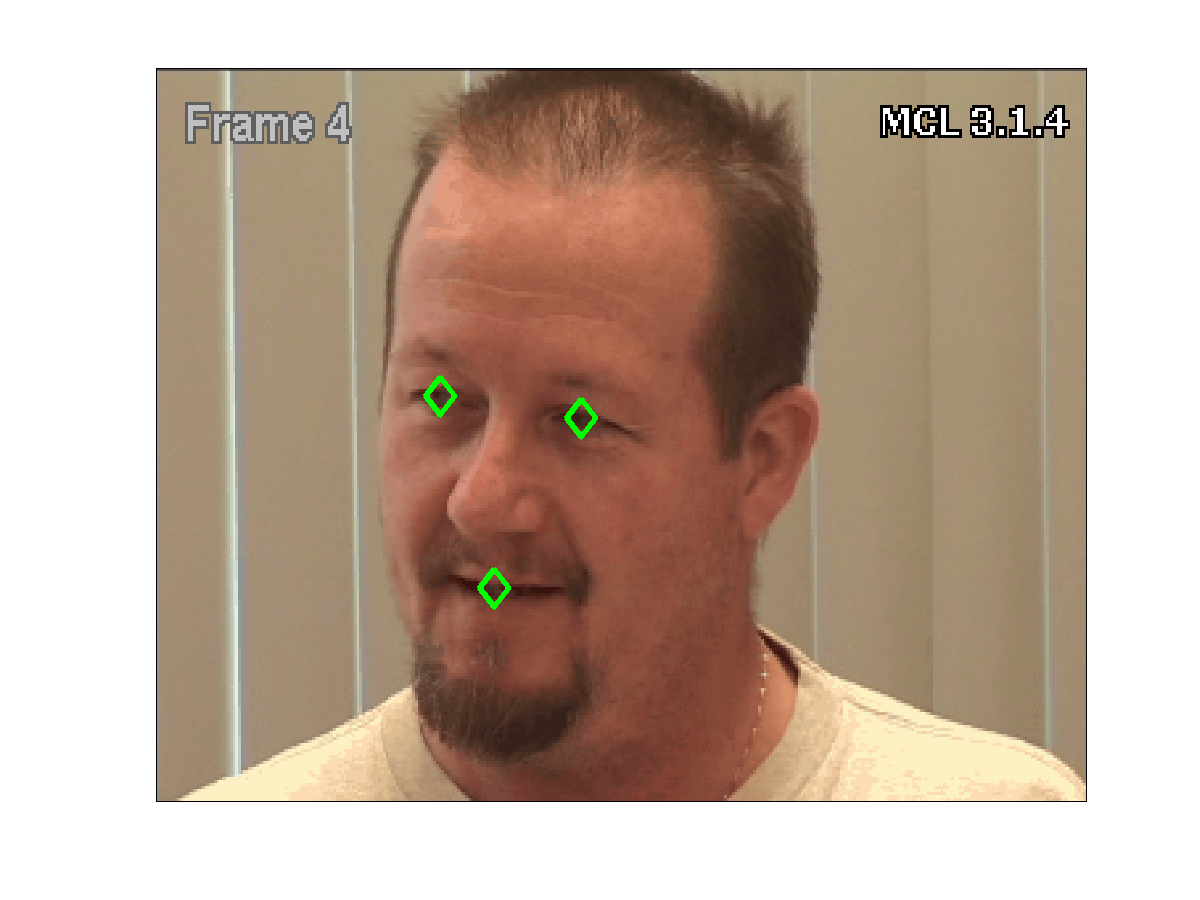};
\end{axis}

\begin{axis}[%
width=0.125\figurewidth,
height=0.419\figureheight,
at={(0.33\figurewidth,0.581\figureheight)},
scale only axis,
axis on top,
separate axis lines,
every outer x axis line/.append style={black},
every x tick label/.append style={font=\color{black}},
xmin=0.5,
xmax=1200.5,
xtick={\empty},
every outer y axis line/.append style={black},
every y tick label/.append style={font=\color{black}},
y dir=reverse,
ymin=0.5,
ymax=900.5,
ytick={\empty},
axis background/.style={fill=white},
title={frame 5}
]
\addplot [forget plot] graphics [xmin=0.5,xmax=1200.5,ymin=0.5,ymax=900.5] {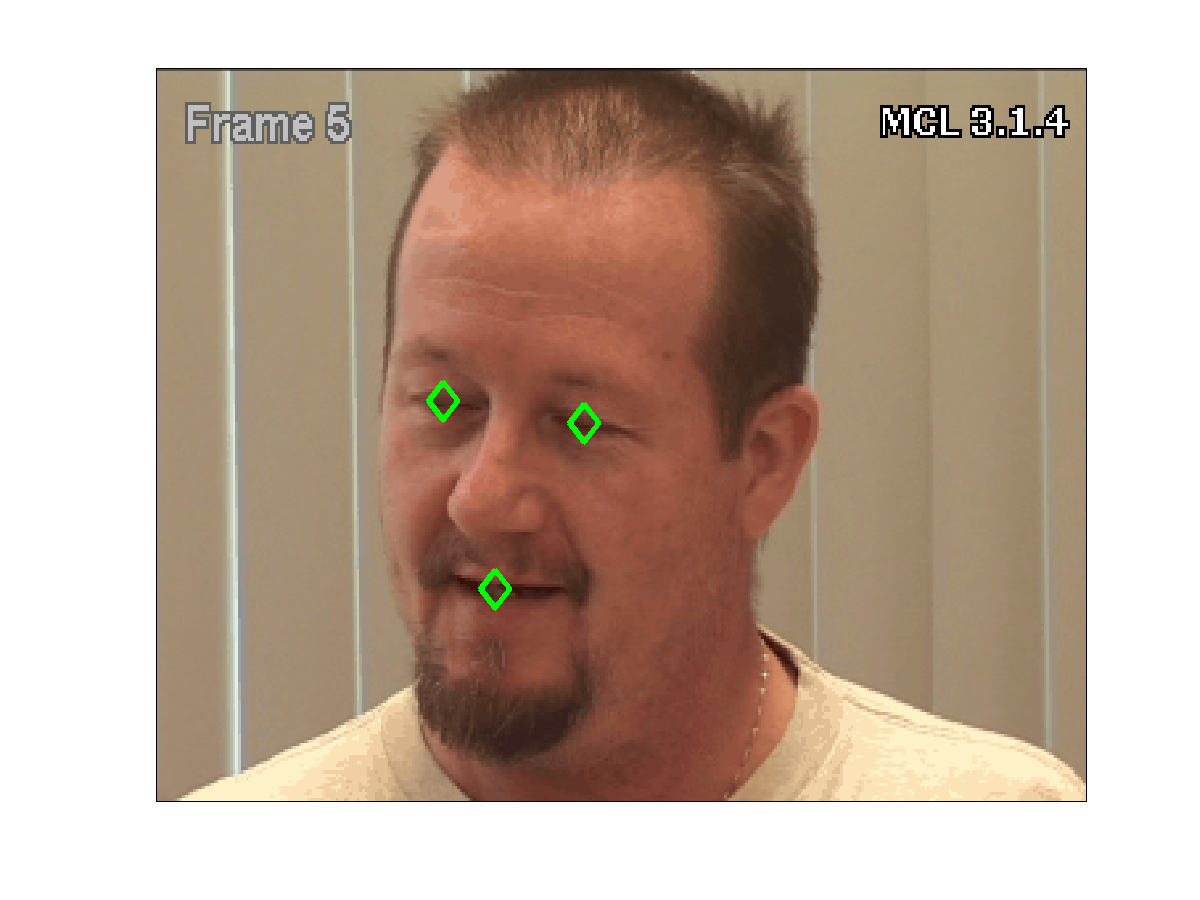};
\end{axis}

\begin{axis}[%
width=0.125\figurewidth,
height=0.419\figureheight,
at={(0.495\figurewidth,0.581\figureheight)},
scale only axis,
axis on top,
separate axis lines,
every outer x axis line/.append style={black},
every x tick label/.append style={font=\color{black}},
xmin=0.5,
xmax=1200.5,
xtick={\empty},
every outer y axis line/.append style={black},
every y tick label/.append style={font=\color{black}},
y dir=reverse,
ymin=0.5,
ymax=900.5,
ytick={\empty},
axis background/.style={fill=white},
title={frame 17}
]
\addplot [forget plot] graphics [xmin=0.5,xmax=1200.5,ymin=0.5,ymax=900.5] {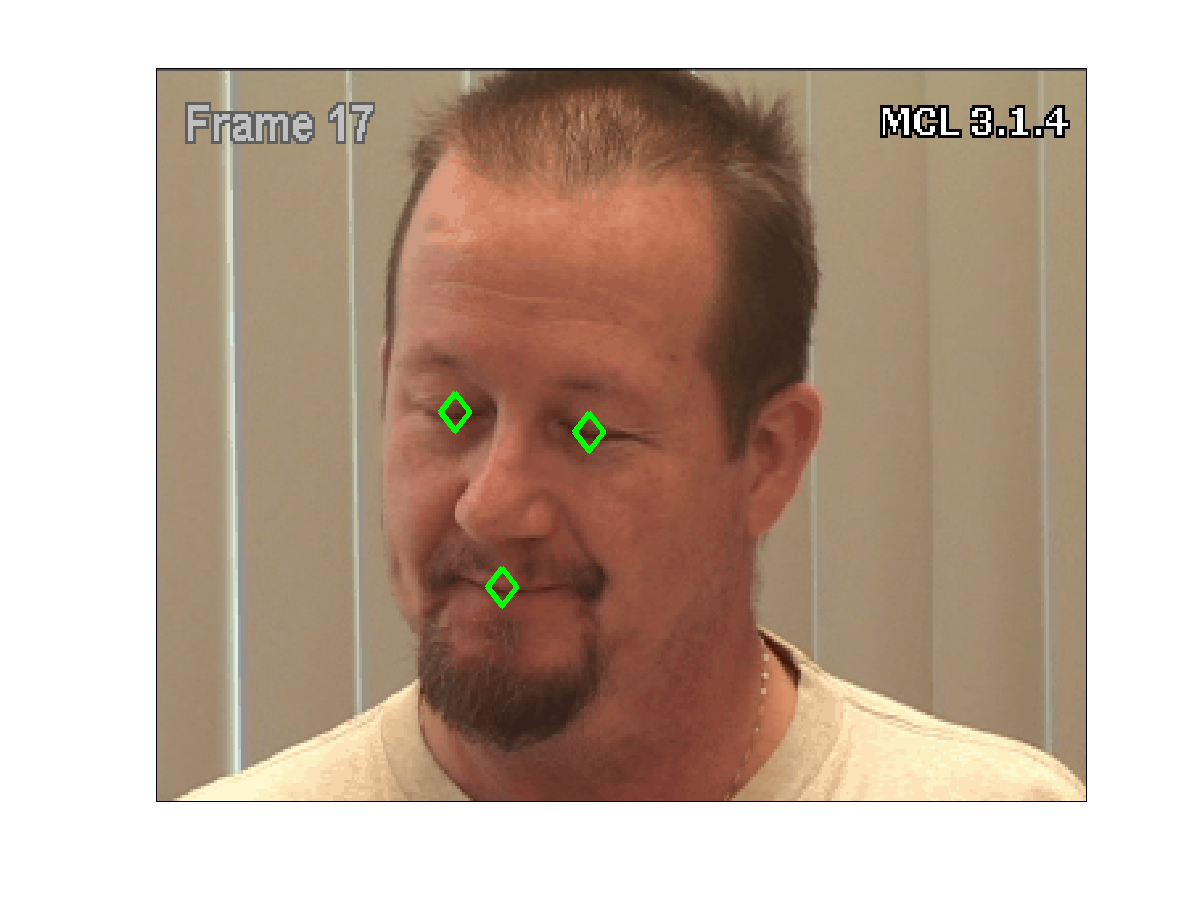};
\end{axis}

\begin{axis}[%
width=0.125\figurewidth,
height=0.419\figureheight,
at={(0.66\figurewidth,0.581\figureheight)},
scale only axis,
axis on top,
separate axis lines,
every outer x axis line/.append style={black},
every x tick label/.append style={font=\color{black}},
xmin=0.5,
xmax=1200.5,
xtick={\empty},
every outer y axis line/.append style={black},
every y tick label/.append style={font=\color{black}},
y dir=reverse,
ymin=0.5,
ymax=900.5,
ytick={\empty},
axis background/.style={fill=white},
title={frame 19}
]
\addplot [forget plot] graphics [xmin=0.5,xmax=1200.5,ymin=0.5,ymax=900.5] {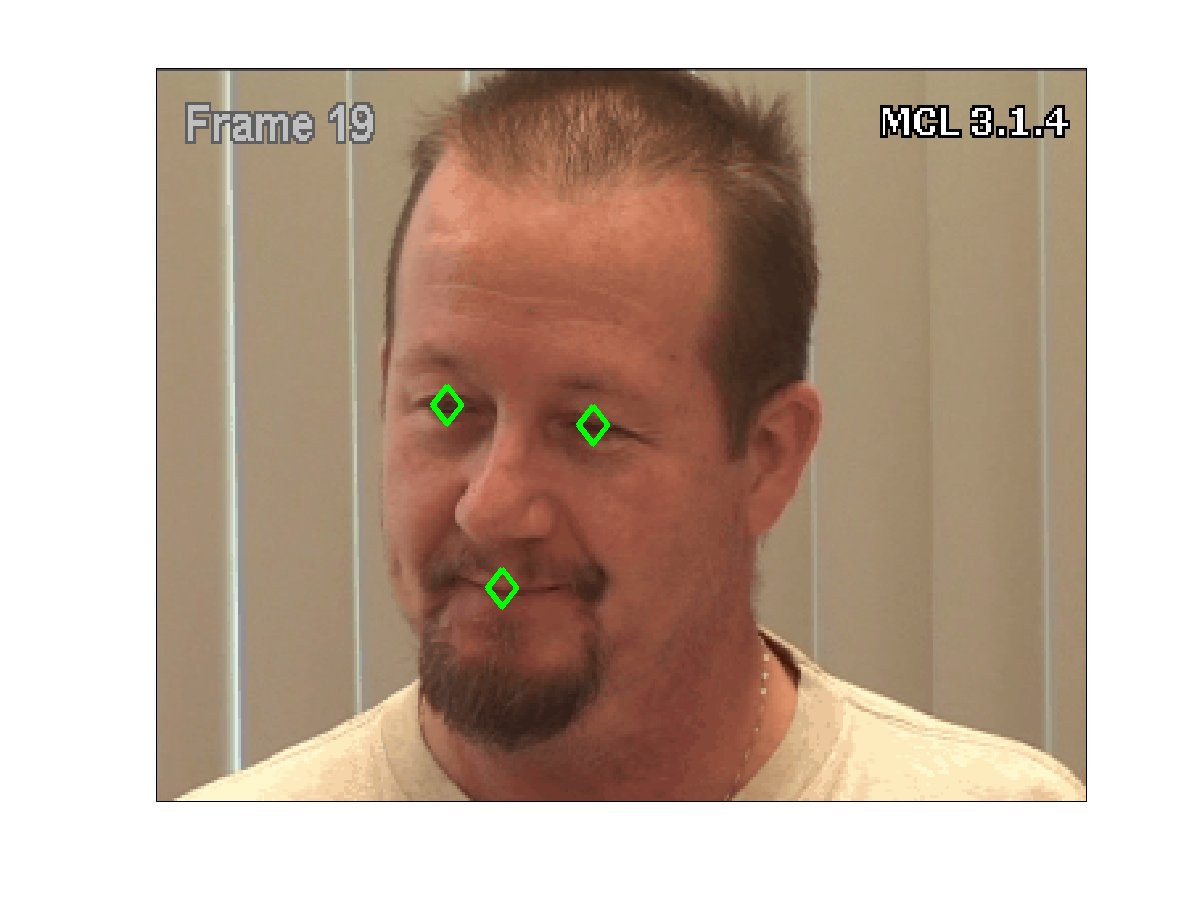};
\end{axis}

\begin{axis}[%
width=0.125\figurewidth,
height=0.419\figureheight,
at={(0.825\figurewidth,0.581\figureheight)},
scale only axis,
axis on top,
separate axis lines,
every outer x axis line/.append style={black},
every x tick label/.append style={font=\color{black}},
xmin=0.5,
xmax=1200.5,
xtick={\empty},
every outer y axis line/.append style={black},
every y tick label/.append style={font=\color{black}},
y dir=reverse,
ymin=0.5,
ymax=900.5,
ytick={\empty},
axis background/.style={fill=white},
title={frame 20}
]
\addplot [forget plot] graphics [xmin=0.5,xmax=1200.5,ymin=0.5,ymax=900.5] {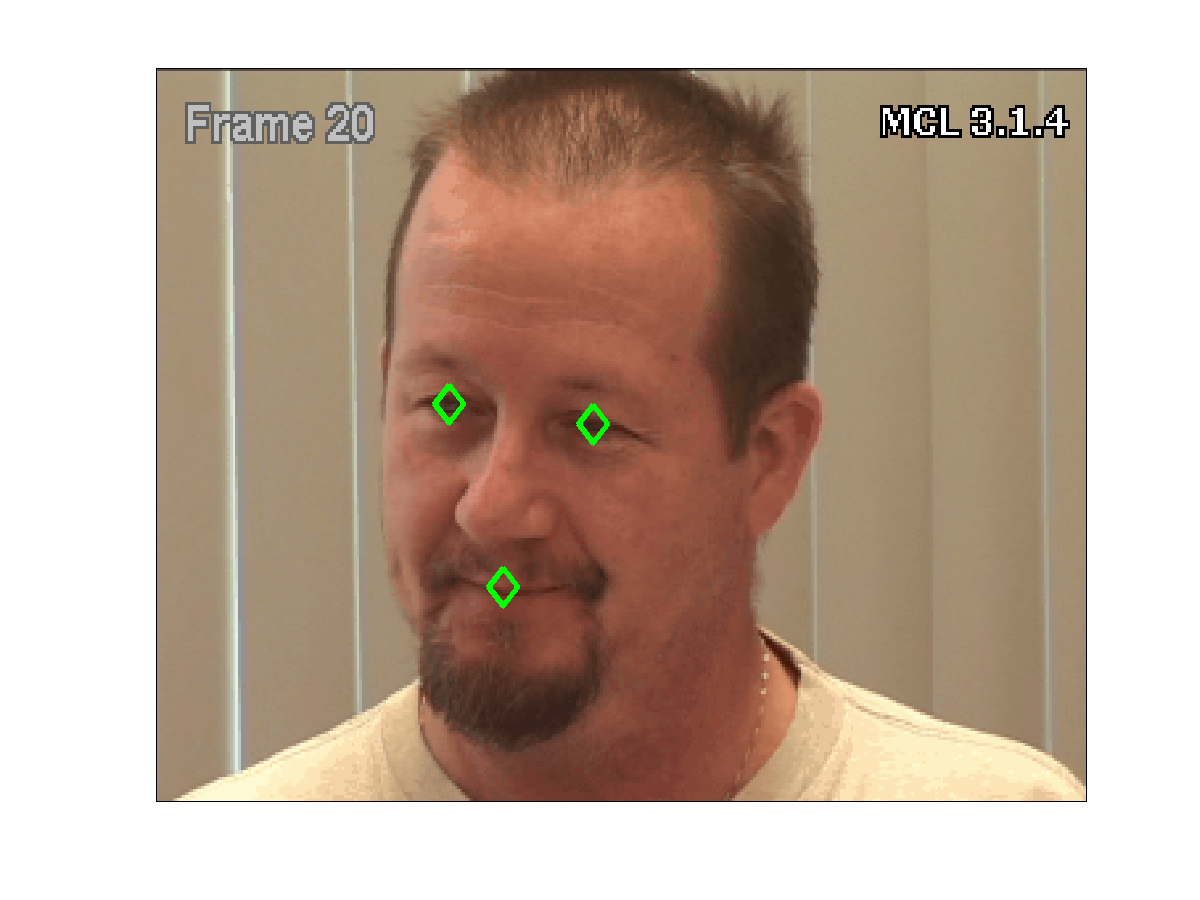};
\end{axis}

\begin{axis}[%
width=0.125\figurewidth,
height=0.419\figureheight,
at={(0\figurewidth,0\figureheight)},
scale only axis,
axis on top,
separate axis lines,
every outer x axis line/.append style={black},
every x tick label/.append style={font=\color{black}},
xmin=0.5,
xmax=1200.5,
xtick={\empty},
every outer y axis line/.append style={black},
every y tick label/.append style={font=\color{black}},
y dir=reverse,
ymin=0.5,
ymax=900.5,
ytick={\empty},
axis background/.style={fill=white}
]
\addplot [forget plot] graphics [xmin=0.5,xmax=1200.5,ymin=0.5,ymax=900.5] {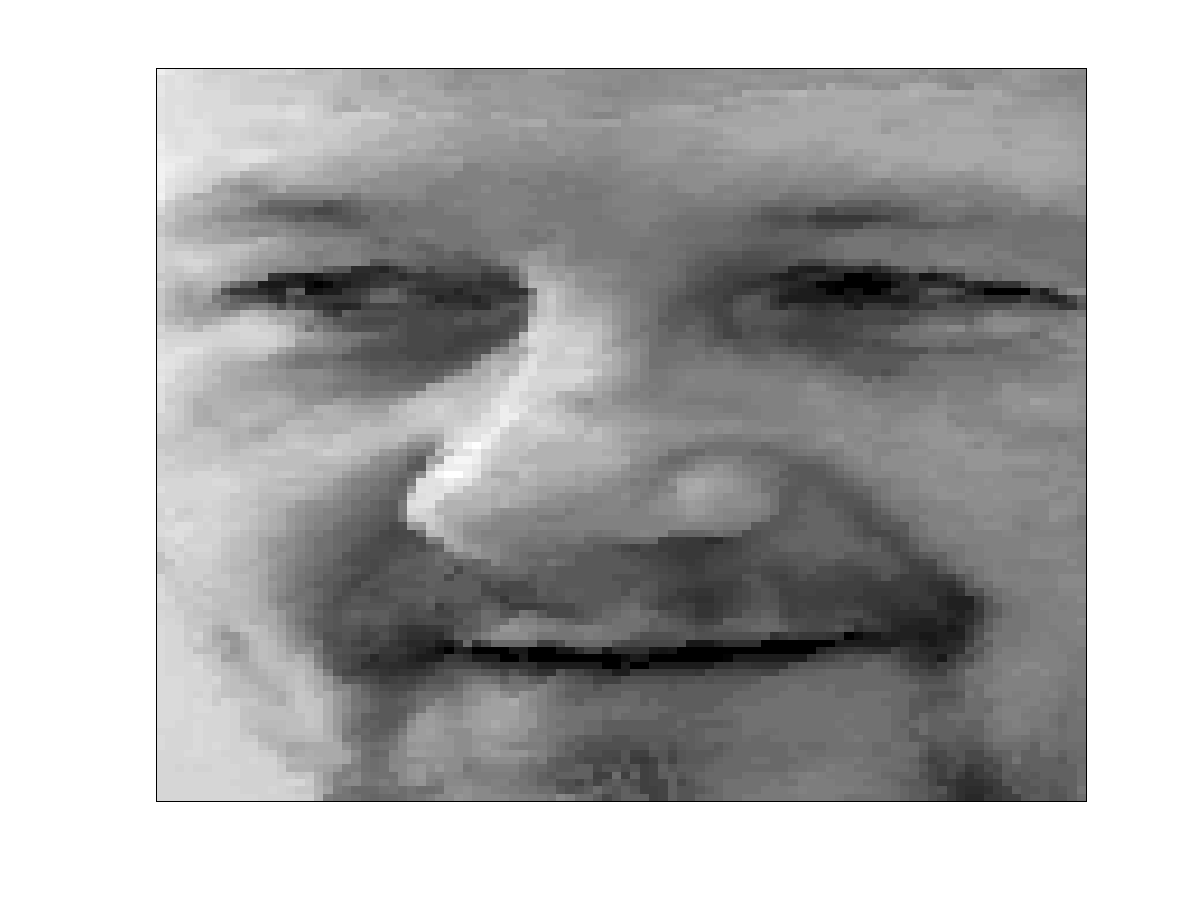};
\end{axis}

\begin{axis}[%
width=0.125\figurewidth,
height=0.419\figureheight,
at={(0.165\figurewidth,0\figureheight)},
scale only axis,
axis on top,
separate axis lines,
every outer x axis line/.append style={black},
every x tick label/.append style={font=\color{black}},
xmin=0.5,
xmax=1200.5,
xtick={\empty},
every outer y axis line/.append style={black},
every y tick label/.append style={font=\color{black}},
y dir=reverse,
ymin=0.5,
ymax=900.5,
ytick={\empty},
axis background/.style={fill=white}
]
\addplot [forget plot] graphics [xmin=0.5,xmax=1200.5,ymin=0.5,ymax=900.5] {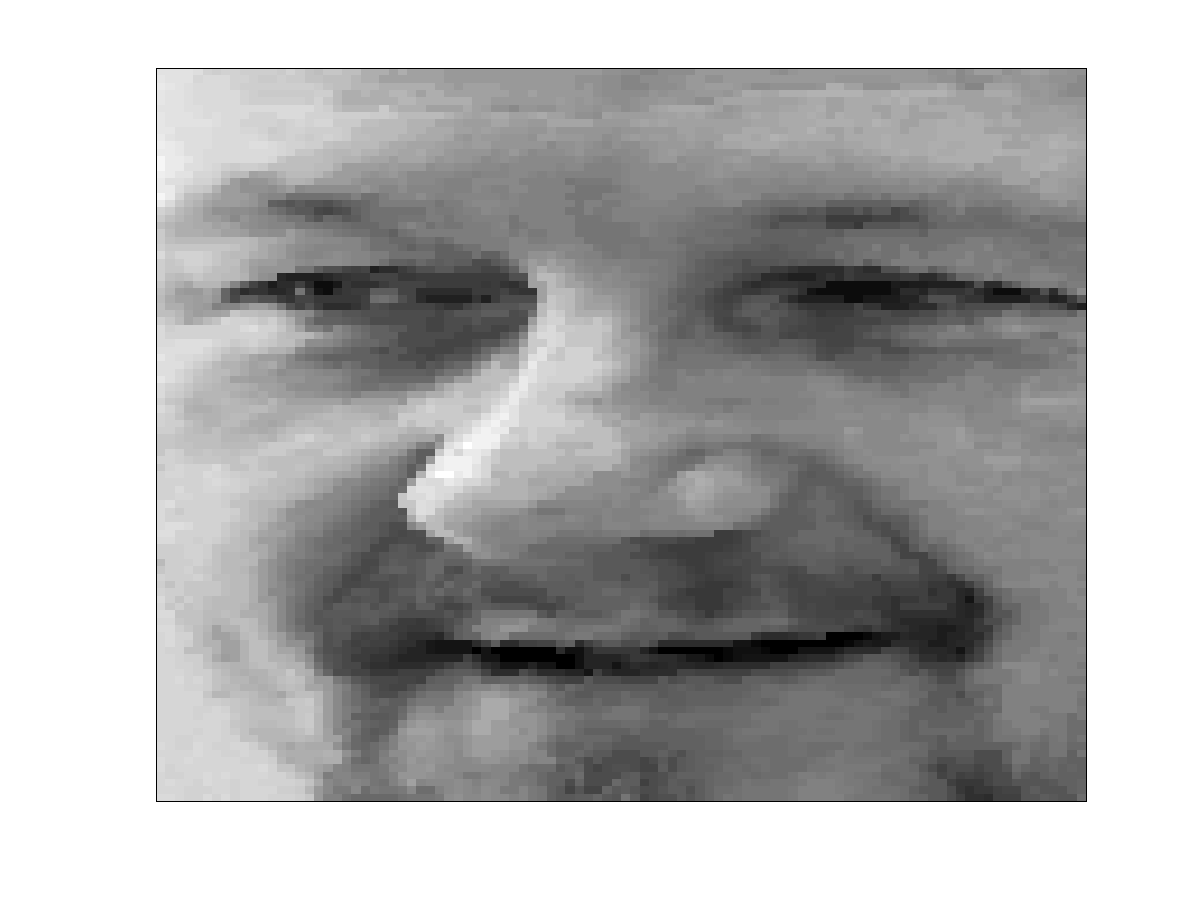};
\end{axis}

\begin{axis}[%
width=0.125\figurewidth,
height=0.419\figureheight,
at={(0.33\figurewidth,0\figureheight)},
scale only axis,
axis on top,
separate axis lines,
every outer x axis line/.append style={black},
every x tick label/.append style={font=\color{black}},
xmin=0.5,
xmax=1200.5,
xtick={\empty},
every outer y axis line/.append style={black},
every y tick label/.append style={font=\color{black}},
y dir=reverse,
ymin=0.5,
ymax=900.5,
ytick={\empty},
axis background/.style={fill=white}
]
\addplot [forget plot] graphics [xmin=0.5,xmax=1200.5,ymin=0.5,ymax=900.5] {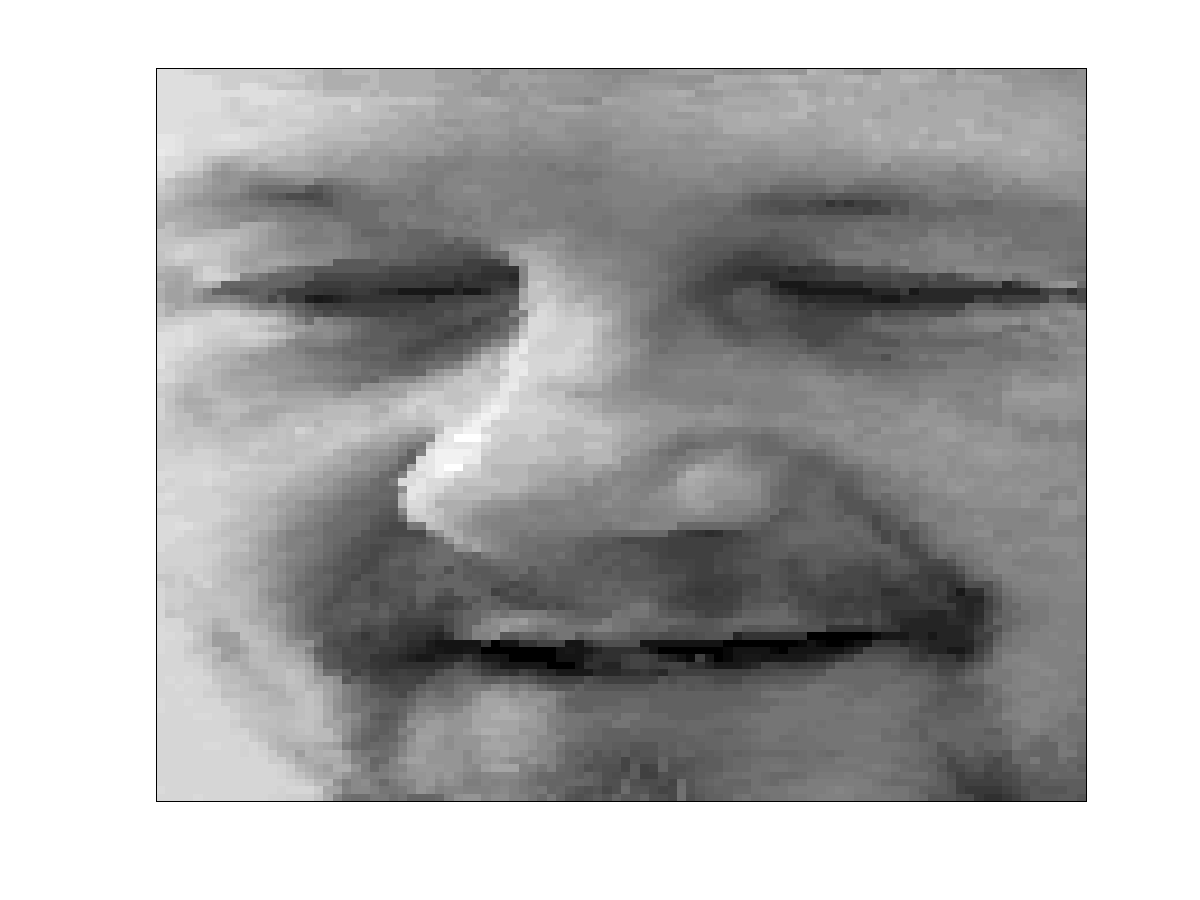};
\end{axis}

\begin{axis}[%
width=0.125\figurewidth,
height=0.419\figureheight,
at={(0.495\figurewidth,0\figureheight)},
scale only axis,
axis on top,
separate axis lines,
every outer x axis line/.append style={black},
every x tick label/.append style={font=\color{black}},
xmin=0.5,
xmax=1200.5,
xtick={\empty},
every outer y axis line/.append style={black},
every y tick label/.append style={font=\color{black}},
y dir=reverse,
ymin=0.5,
ymax=900.5,
ytick={\empty},
axis background/.style={fill=white}
]
\addplot [forget plot] graphics [xmin=0.5,xmax=1200.5,ymin=0.5,ymax=900.5] {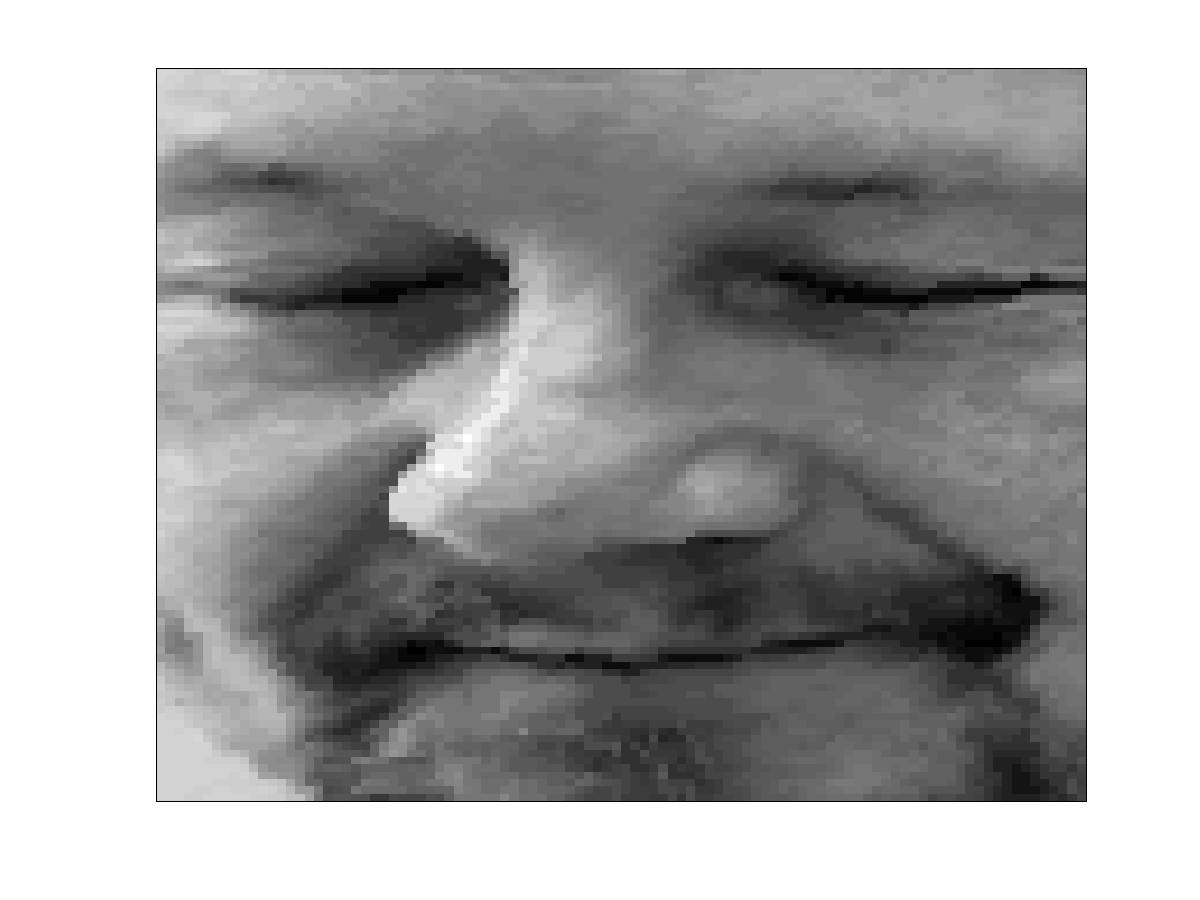};
\end{axis}

\begin{axis}[%
width=0.125\figurewidth,
height=0.419\figureheight,
at={(0.66\figurewidth,0\figureheight)},
scale only axis,
axis on top,
separate axis lines,
every outer x axis line/.append style={black},
every x tick label/.append style={font=\color{black}},
xmin=0.5,
xmax=1200.5,
xtick={\empty},
every outer y axis line/.append style={black},
every y tick label/.append style={font=\color{black}},
y dir=reverse,
ymin=0.5,
ymax=900.5,
ytick={\empty},
axis background/.style={fill=white}
]
\addplot [forget plot] graphics [xmin=0.5,xmax=1200.5,ymin=0.5,ymax=900.5] {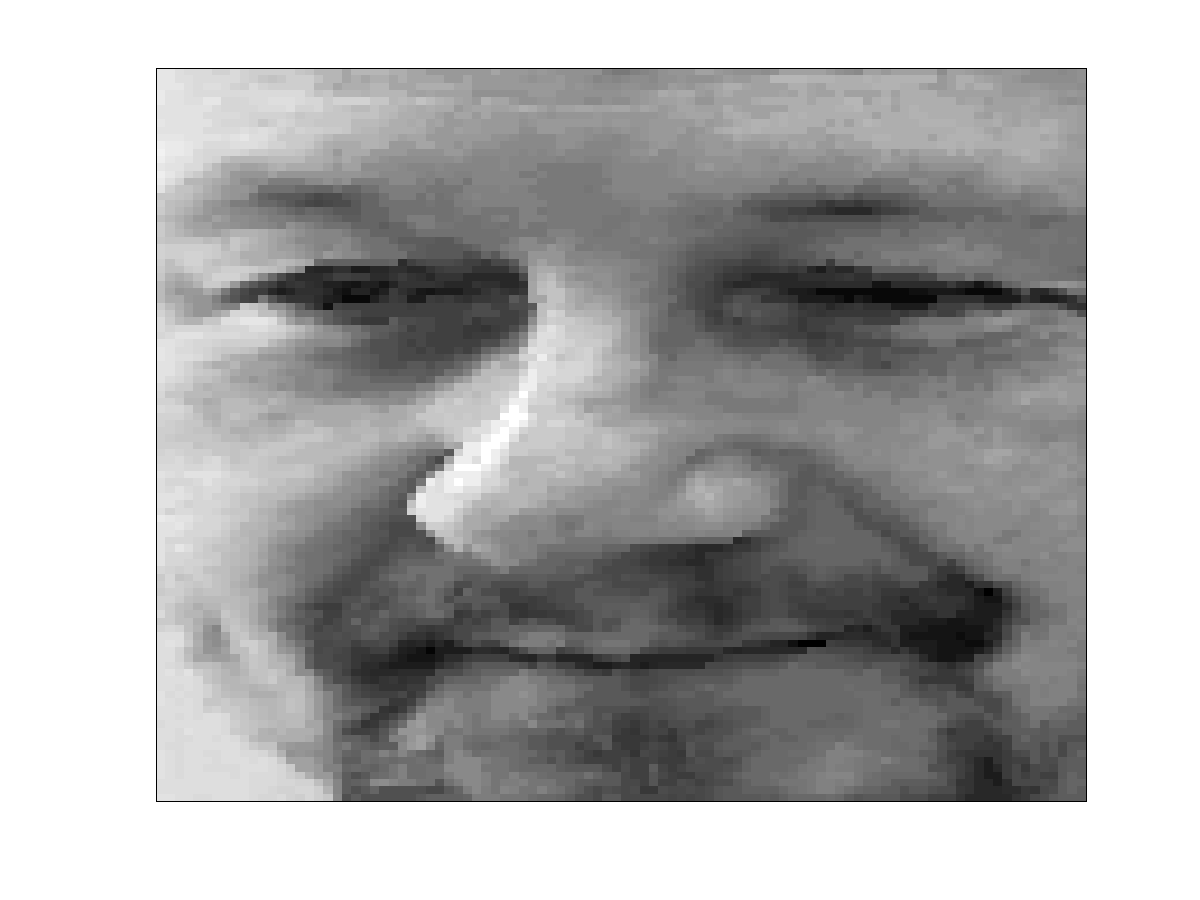};
\end{axis}

\begin{axis}[%
width=0.125\figurewidth,
height=0.419\figureheight,
at={(0.825\figurewidth,0\figureheight)},
scale only axis,
axis on top,
separate axis lines,
every outer x axis line/.append style={black},
every x tick label/.append style={font=\color{black}},
xmin=0.5,
xmax=1200.5,
xtick={\empty},
every outer y axis line/.append style={black},
every y tick label/.append style={font=\color{black}},
y dir=reverse,
ymin=0.5,
ymax=900.5,
ytick={\empty},
axis background/.style={fill=white}
]
\addplot [forget plot] graphics [xmin=0.5,xmax=1200.5,ymin=0.5,ymax=900.5] {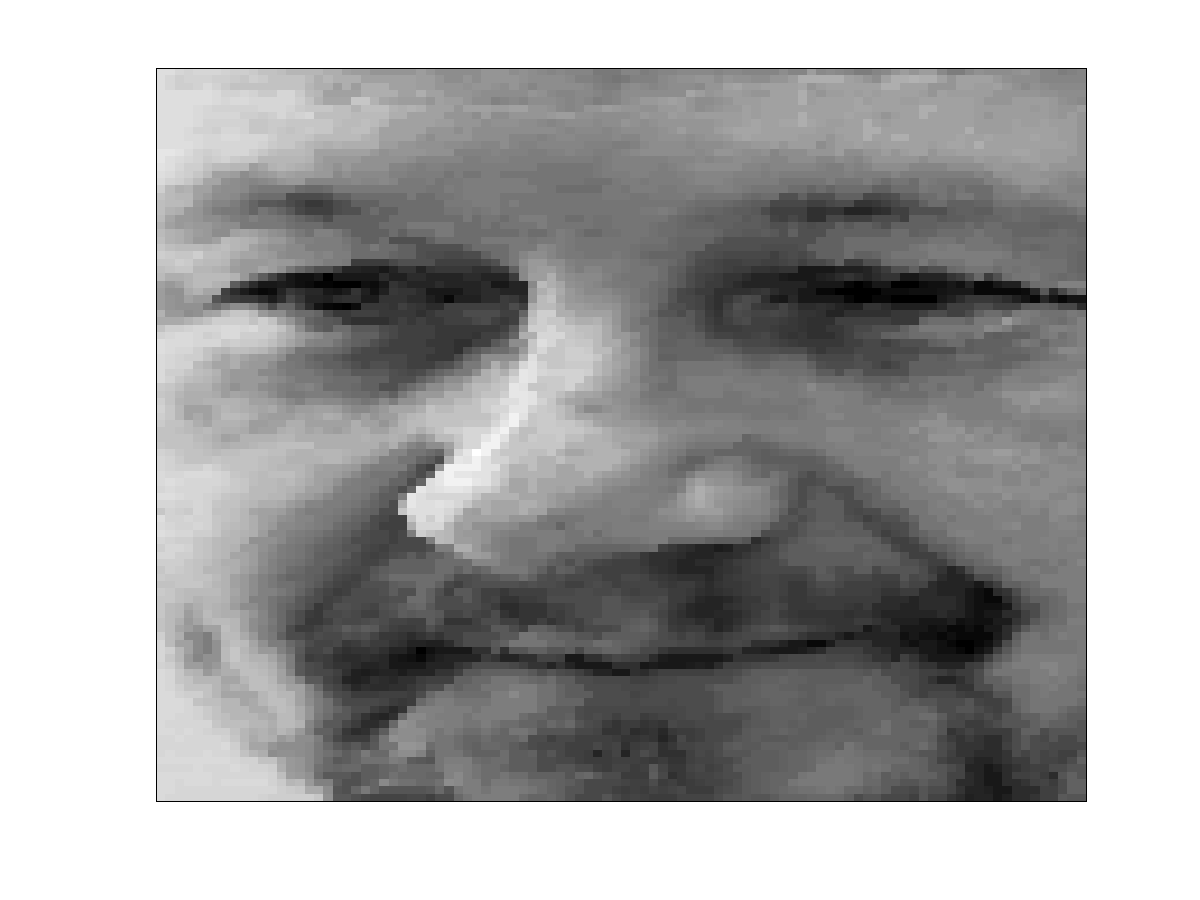};
\end{axis}
\end{tikzpicture}%

%% file: time.tikz
%
%
\begin{tikzpicture}

\begin{axis}[%
width=0.951\figurewidth,
height=\figureheight,
at={(0\figurewidth,0\figureheight)},
scale only axis,
separate axis lines,
every outer x axis line/.append style={black},
every x tick label/.append style={font=\color{black}},
xmin=0,
xmax=25,
xlabel={Frame},
xmajorgrids,
every outer y axis line/.append style={black},
every y tick label/.append style={font=\color{black}},
ymin=-0.8,
ymax=-0.1,
ylabel={Gray level avg},
ymajorgrids,
axis background/.style={fill=white},
title={Left eye region}
]
\addplot [color=blue,solid,line width=2.0pt,forget plot]
  table[row sep=crcr]{%
1	-0.220754525984161\\
2	-0.207361596500984\\
3	-0.198141156042755\\
4	-0.192007615522481\\
5	-0.262926972257285\\
6	-0.423622871659726\\
7	-0.559492612653224\\
8	-0.689682653773941\\
9	-0.690092477282116\\
10	-0.651763873776747\\
11	-0.746795824277418\\
12	-0.708168657990205\\
13	-0.716927712825269\\
14	-0.692606013281907\\
15	-0.655372665143721\\
16	-0.631352080109302\\
17	-0.376445982328181\\
18	-0.370892841071387\\
19	-0.315084932708182\\
20	-0.137014529773995\\
21	-0.140331190391939\\
22	-0.129392493435331\\
23	-0.167953464712891\\
24	-0.193349291149835\\
25	-0.206636106102794\\
};
\end{axis}
\end{tikzpicture}%

%% file: timeAll.tikz
%
%
\begin{tikzpicture}

\begin{axis}[%
width=0.951\figurewidth,
height=\figureheight,
at={(0\figurewidth,0\figureheight)},
scale only axis,
separate axis lines,
every outer x axis line/.append style={black},
every x tick label/.append style={font=\color{black}},
xmin=0,
xmax=160,
xlabel={Frame},
xmajorgrids,
every outer y axis line/.append style={black},
every y tick label/.append style={font=\color{black}},
ymin=-1,
ymax=-0.1,
ylabel={Gray level avg},
ymajorgrids,
axis background/.style={fill=white},
title={Left eye region},
legend style={legend cell align=left,align=left,draw=black}
]
\addplot [color=blue,solid,line width=2.0pt]
  table[row sep=crcr]{%
1	-0.220754525984161\\
2	-0.207361596500984\\
3	-0.198141156042755\\
4	-0.192007615522481\\
5	-0.262926972257285\\
6	-0.423622871659726\\
7	-0.559492612653224\\
8	-0.689682653773941\\
9	-0.690092477282116\\
10	-0.651763873776747\\
11	-0.746795824277418\\
12	-0.708168657990205\\
13	-0.716927712825269\\
14	-0.692606013281907\\
15	-0.655372665143721\\
16	-0.631352080109302\\
17	-0.376445982328181\\
18	-0.370892841071387\\
19	-0.315084932708182\\
20	-0.137014529773995\\
21	-0.140331190391939\\
22	-0.129392493435331\\
23	-0.167953464712891\\
24	-0.193349291149835\\
25	-0.206636106102794\\
26	-0.204564623601127\\
27	-0.164628194063398\\
28	-0.199537874248269\\
29	-0.182083844962223\\
30	-0.197016677020127\\
31	-0.152643256908414\\
32	-0.202344828347043\\
33	-0.198563752273903\\
34	-0.221724154238796\\
35	-0.231157946411941\\
36	-0.208109225770864\\
37	-0.283273321855503\\
38	-0.185322868618846\\
39	-0.247244786294442\\
40	-0.243628528776993\\
41	-0.223776327686022\\
42	-0.239522607497626\\
43	-0.213511389254377\\
44	-0.23859135051985\\
45	-0.217653012268548\\
46	-0.208853669806813\\
47	-0.279491878209842\\
48	-0.2961625782213\\
49	-0.32041234987932\\
50	-0.29937740643725\\
51	-0.306115717165722\\
52	-0.277253332067867\\
53	-0.278867641039543\\
54	-0.286354543176567\\
55	-0.273601612761948\\
56	-0.26889631325399\\
57	-0.280320760837345\\
58	-0.281586055645781\\
59	-0.261764693799509\\
60	-0.300027918656029\\
61	-0.308387451265149\\
62	-0.29633287172906\\
63	-0.30368128439554\\
64	-0.264654416422357\\
65	-0.311129976029727\\
66	-0.33091299355511\\
67	-0.299194006429965\\
68	-0.330985502983418\\
69	-0.335118217205687\\
70	-0.291154160831016\\
71	-0.279046154466507\\
72	-0.324329881089191\\
73	-0.305330846551686\\
74	-0.297599392715823\\
75	-0.338612515936735\\
76	-0.302486518693718\\
77	-0.323083391326646\\
78	-0.337242195421957\\
79	-0.266697778933611\\
80	-0.315286313961625\\
81	-0.287527456159066\\
82	-0.334439152606725\\
83	-0.299397742308941\\
84	-0.315125961956595\\
85	-0.354028483156877\\
86	-0.76409185397372\\
87	-0.837358841840531\\
88	-0.822057955793044\\
89	-0.83735989479384\\
90	-0.81122481545232\\
91	-0.812747344142725\\
92	-0.822142354070873\\
93	-0.778770543142139\\
94	-0.793246235960194\\
95	-0.789825897435391\\
96	-0.813249122464643\\
97	-0.90814338140327\\
98	-0.768228648256685\\
99	-0.800904401053711\\
100	-0.873421901323608\\
101	-0.793386028382865\\
102	-0.780102780393237\\
103	-0.813013739627951\\
104	-0.834364966719947\\
105	-0.793506940724111\\
106	-0.82742567551005\\
107	-0.803005774020167\\
108	-0.855985663249588\\
109	-0.870810209916814\\
110	-0.84476407070634\\
111	-0.839310295053065\\
112	-0.773192273760197\\
113	-0.781334718424246\\
114	-0.765239993097493\\
115	-0.795934585581629\\
116	-0.828204125484501\\
117	-0.818426176248517\\
118	-0.709374500790762\\
119	-0.529879554071715\\
120	-0.498111043743069\\
121	-0.464161964131316\\
122	-0.471515687798995\\
123	-0.387365985505386\\
124	-0.388354780162586\\
125	-0.380310090567395\\
126	-0.326713983320927\\
127	-0.365748773934927\\
128	-0.41160271872447\\
129	-0.398018948510381\\
130	-0.443606921035135\\
131	-0.425454449825529\\
132	-0.445210128959842\\
133	-0.39541773326415\\
134	-0.401324725748662\\
135	-0.39167549097542\\
};
\addlegendentry{Pain};

\addplot [color=red,solid,line width=2.0pt]
  table[row sep=crcr]{%
1	-0.372879507677546\\
2	-0.35017156875416\\
3	-0.320290281334413\\
4	-0.402367451843466\\
5	-0.357157827078137\\
6	-0.337281556185164\\
7	-0.332153949644884\\
8	-0.335390024382273\\
9	-0.356540469897583\\
10	-0.337049624267606\\
11	-0.385553344843633\\
12	-0.320987663191358\\
13	-0.270233460714088\\
14	-0.235956535387423\\
15	-0.205291489939748\\
16	-0.210183046496867\\
17	-0.221725926458749\\
18	-0.216308007081986\\
19	-0.252483453730349\\
20	-0.230104754285279\\
21	-0.209232096992253\\
22	-0.219946623399187\\
23	-0.206037252924182\\
24	-0.209062109130966\\
25	-0.184398818817948\\
26	-0.195245440069402\\
27	-0.208289597457393\\
28	-0.225173926886956\\
29	-0.229211285993624\\
30	-0.268596992457079\\
31	-0.254580105207544\\
32	-0.306202774409846\\
33	-0.259983506767455\\
34	-0.257927219925621\\
35	-0.300336756740179\\
36	-0.288843637206138\\
37	-0.25739787402948\\
38	-0.239303641554336\\
39	-0.301480397321417\\
40	-0.275921756154237\\
41	-0.278306423108414\\
42	-0.307484079517776\\
43	-0.227210775643943\\
44	-0.287109140289627\\
45	-0.254905255684389\\
46	-0.219144247863432\\
47	-0.179883476389658\\
48	-0.233284587390911\\
49	-0.23895954543291\\
50	-0.259569582455823\\
51	-0.245797327441128\\
52	-0.213899039366529\\
53	-0.234334628997775\\
54	-0.241100327846994\\
55	-0.248559801536979\\
56	-0.279257644196315\\
57	-0.228454719389233\\
58	-0.275232250645269\\
59	-0.253946682170304\\
60	-0.246325892672713\\
61	-0.280689490268549\\
62	-0.294256890529781\\
63	-0.252860247346507\\
64	-0.274562244018573\\
65	-0.259456802382425\\
66	-0.26169510817999\\
67	-0.289753600414388\\
68	-0.269527226279735\\
69	-0.243956470269286\\
70	-0.246491418274433\\
71	-0.234859990800976\\
72	-0.267230208991255\\
73	-0.26278944749703\\
74	-0.211730305708688\\
75	-0.243546715500889\\
76	-0.256210291324376\\
77	-0.272490823946664\\
78	-0.214570864869956\\
79	-0.265841791450226\\
80	-0.240921490614489\\
81	-0.275499523195111\\
82	-0.258541412897451\\
83	-0.251289970556271\\
84	-0.207287864086676\\
85	-0.16500007256365\\
86	-0.120653255400539\\
87	-0.198061958778432\\
88	-0.144952883704337\\
89	-0.161790279596324\\
90	-0.168985050355329\\
91	-0.176288590113702\\
92	-0.186026237646277\\
93	-0.167925739749191\\
94	-0.156860847566008\\
95	-0.180800170402847\\
96	-0.191372181205061\\
97	-0.16176857480147\\
98	-0.119058177565857\\
99	-0.160912932283489\\
100	-0.215430055159734\\
101	-0.174478746499106\\
102	-0.181951145840279\\
103	-0.142937969874124\\
104	-0.158721368700937\\
105	-0.16831131508645\\
106	-0.164855953703772\\
107	-0.162579177972173\\
108	-0.145355144136441\\
109	-0.175668500182059\\
110	-0.151496942425073\\
111	-0.176314891401958\\
112	-0.173852921091843\\
113	-0.193858697045101\\
114	-0.225363542430639\\
115	-0.38436814645253\\
116	-0.270508502011248\\
117	-0.201371921845761\\
118	-0.190511005928997\\
119	-0.212440080171393\\
120	-0.241077238326954\\
121	-0.193834630848728\\
122	-0.196488133530922\\
123	-0.151121247607033\\
124	-0.223543107247243\\
125	-0.235282490600404\\
126	-0.25133384153311\\
127	-0.234118597895852\\
128	-0.273185666007889\\
129	-0.257198184995772\\
130	-0.309452145778319\\
131	-0.301079056182261\\
132	-0.291725062992864\\
133	-0.278554846974151\\
134	-0.320768554871647\\
135	-0.296968858206694\\
136	-0.297232257406726\\
137	-0.304615233325883\\
138	-0.25255056458372\\
139	-0.291692118562116\\
140	-0.300829375674011\\
141	-0.332814246725351\\
142	-0.323250638500061\\
143	-0.297714401638071\\
144	-0.337377677179243\\
145	-0.297662857212994\\
146	-0.314366998049052\\
147	-0.309003602990334\\
148	-0.288591086585299\\
149	-0.275824380773047\\
150	-0.281746402888783\\
151	-0.263794489860956\\
152	-0.261176035238723\\
153	-0.25321889447766\\
154	-0.235653121100778\\
155	-0.293914382681658\\
156	-0.28027146107877\\
157	-0.269096080428507\\
};
\addlegendentry{No pain};

\end{axis}
\end{tikzpicture}%

%% file: filterBankRes.tikz
%
%
\begin{tikzpicture}

\begin{axis}[%
width=0.385\figurewidth,
height=\figureheight,
at={(0\figurewidth,0\figureheight)},
scale only axis,
every outer x axis line/.append style={black},
every x tick label/.append style={font=\color{black}},
xmin=0,
xmax=150,
xlabel={Frame},
xmajorgrids,
every outer y axis line/.append style={black},
every y tick label/.append style={font=\color{black}},
ymin=-0.06,
ymax=0.06,
ylabel={Value},
ymajorgrids,
axis background/.style={fill=white},
title={Coefficients - Scale 15},
axis x line*=bottom,
axis y line*=left
]
\addplot [color=blue,solid,line width=2.0pt,forget plot]
  table[row sep=crcr]{%
1	-0.0592870918217354\\
2	-0.0228790359872223\\
3	-0.023005384921741\\
4	-0.0144292063770058\\
5	0.0146703906080824\\
6	0.0234046477764475\\
7	0.0389320631758711\\
8	0.0359399061978003\\
9	0.0367092454184592\\
10	0.0321413460232101\\
11	0.0374295269840791\\
12	0.041228991824069\\
13	0.0354242235722072\\
14	0.0378019782584874\\
15	0.0319328215581344\\
16	0.0356612824066192\\
17	0.0167498802562624\\
18	0.0179951943163254\\
19	0.0128032649406607\\
20	-0.00172826377417581\\
21	0.00201523040106217\\
22	-0.0106600545471872\\
23	0.00507835790098292\\
24	-0.000985715884165533\\
25	-0.000467955912616221\\
26	0.00172721876553821\\
27	-0.0022730840063006\\
28	0.00360674342568647\\
29	-0.00222227748519785\\
30	0.00411590474260392\\
31	0.00288822457066241\\
32	-0.00402040375623868\\
33	-0.0060053413182105\\
34	-0.0102964086064991\\
35	-0.00376445936218873\\
36	-0.00488061433369797\\
37	0.00186404427664897\\
38	-0.00276179403178825\\
39	0.000643594476466427\\
40	-0.000828035727725879\\
41	-0.00233134295703006\\
42	0.000214064460808524\\
43	-0.00295532503236666\\
44	0.000844090375776968\\
45	-0.0029193539404516\\
46	-0.0070289809353178\\
47	0.000660635071891188\\
48	0.00173614699014514\\
49	0.00385341973121441\\
50	-0.000506415444038677\\
51	-0.00454552091902885\\
52	0.000886964787644129\\
53	-0.00558037484122338\\
54	-0.00458372861854751\\
55	0.00302145161711957\\
56	0.031659723886313\\
57	0.0312679893666329\\
58	0.0328654878890302\\
59	0.0335950535441135\\
60	0.0289351801489919\\
61	0.0344107838680292\\
62	0.0314872295431094\\
63	0.0271978291129177\\
64	0.034967006301022\\
65	0.0313588830361246\\
66	0.0379404802467748\\
67	0.0358972721746524\\
68	0.0333612444414814\\
69	0.0337180462897472\\
70	0.0304346063893913\\
71	-0.0303834350065378\\
72	-0.0380190014799089\\
73	-0.0350514216937633\\
74	-0.0361836953434607\\
75	-0.0326886782829196\\
76	-0.0330388329387788\\
77	-0.0345463695196622\\
78	-0.0242875485075155\\
79	-0.0299322988713309\\
80	-0.0279734273468545\\
81	-0.0333106995811436\\
82	-0.0472436890959746\\
83	-0.0303816557186789\\
84	-0.0347628564702222\\
85	-0.0397920489272473\\
86	0.000368261512832698\\
87	0.00637196382017165\\
88	-0.00630633484480642\\
89	-0.0200993656382893\\
90	-0.0185118681501889\\
};
\addplot [color=red,solid,line width=2.0pt,forget plot]
  table[row sep=crcr]{%
1	0.0138062346594237\\
2	0.0141948326831005\\
3	0.00984385159585975\\
4	0.010355184287226\\
5	0.0131523383498505\\
6	0.0138440666271198\\
7	0.00997723845839927\\
8	0.0108412773392164\\
9	0.0159931099304712\\
10	0.0162782495190632\\
11	0.0182245925208829\\
12	0.0141261698529565\\
13	0.00313975883894121\\
14	0.00430954024598687\\
15	-0.0051331492860014\\
16	-0.00532219440365266\\
17	-0.0140530763980857\\
18	-0.00469162793746759\\
19	-0.00162742937919897\\
20	-0.0073999451159504\\
21	-0.00817718999859303\\
22	-0.00539667235288288\\
23	-0.00254902674578104\\
24	-0.0101865571776582\\
25	-0.00792565946356987\\
26	-0.0054739841300741\\
27	-0.0118815894792617\\
28	0.00306564174962264\\
29	-0.00607068749435506\\
30	0.000340824917400893\\
31	0.00646540664994868\\
32	0.0160461808106875\\
33	0.00308497195547602\\
34	0.00363802487189168\\
35	0.00271029294739719\\
36	0.00392960603359153\\
37	0.00795689304738737\\
38	0.00267744065590143\\
39	0.00421574745978102\\
40	0.00168623809031409\\
41	-0.00302325829888259\\
42	0.00785365664870429\\
43	-0.00403095187663765\\
44	-0.000603594556152863\\
45	0.000386679055990202\\
46	-0.0057349627566194\\
47	-0.00907596538154936\\
48	-0.00385766949547645\\
49	-0.00295487674360074\\
50	-0.00122816877963585\\
51	-0.000139557714916126\\
52	-0.00713778323765309\\
53	-0.00356199020036165\\
54	-0.00263498324032685\\
55	-0.00529486416321572\\
56	-0.00465393880033992\\
57	-0.00719624932098958\\
58	-0.0070262507096303\\
59	-0.000514909976716549\\
60	-0.00478549919824912\\
61	-0.00369616681776668\\
62	-0.00431323464781801\\
63	-0.000557049509614312\\
64	-0.0066840327543914\\
65	-0.00277240056291381\\
66	-0.00652878380034476\\
67	-0.00437071003860298\\
68	-0.00759963581779666\\
69	-0.00064708837470514\\
70	0.00879475522045774\\
71	0.0112021484332669\\
72	0.00353716248497796\\
73	0.00772144333083197\\
74	0.00312474101446512\\
75	0.00492586199177872\\
76	0.0045659376533829\\
77	0.00420116844175224\\
78	0.00160496863386774\\
79	0.00851923976668467\\
80	0.00205453948225797\\
81	0.00460467014579653\\
82	0.00725714562575696\\
83	0.0138021541646439\\
84	0.00738836946335585\\
85	0.00790054057978084\\
86	0.00281361762757172\\
87	0.00236877259624229\\
88	0.00330586332567244\\
89	0.00378584149105624\\
90	0.00489597723395896\\
91	0.00269408757032581\\
92	0.00382373434885682\\
93	0.00188911327222273\\
94	0.00193779696327546\\
95	0.00753925174354027\\
96	0.00600508266228365\\
97	0.0032120887009091\\
98	0.000301763298902889\\
99	-0.00217439783880113\\
100	-0.0162569394644671\\
101	-0.0043639467560753\\
102	0.00472882434277479\\
103	0.00269805366601873\\
104	0.00364065088198648\\
105	-0.00112495355738438\\
106	0.0049612632942027\\
107	0.00494787628241413\\
108	0.00637754756707301\\
109	0.00135162694997934\\
110	-0.00121591087344823\\
111	0.000430763670739241\\
112	0.00192442425334665\\
};
\end{axis}

\begin{axis}[%
width=0.385\figurewidth,
height=\figureheight,
at={(0.54\figurewidth,0\figureheight)},
scale only axis,
every outer x axis line/.append style={black},
every x tick label/.append style={font=\color{black}},
xmin=0,
xmax=100,
xlabel={Frame},
xmajorgrids,
every outer y axis line/.append style={black},
every y tick label/.append style={font=\color{black}},
ymin=-0.03,
ymax=0.04,
ylabel={Value},
ymajorgrids,
axis background/.style={fill=white},
title={Coefficients - Scale 20},
axis x line*=bottom,
axis y line*=left
]
\addplot [color=blue,solid,line width=2.0pt,forget plot]
  table[row sep=crcr]{%
1	0.00819342364431522\\
2	0.00940496085639745\\
3	0.00378728079356742\\
4	0.00219501918713305\\
5	0.00336538861601229\\
6	0.0111673647132143\\
7	0.0254864051368135\\
8	0.0293384741749352\\
9	0.0323168568618495\\
10	0.0278553963086872\\
11	0.0373812513813156\\
12	0.0290366166681992\\
13	0.0299333924658503\\
14	0.0267756123990441\\
15	0.0233329192540894\\
16	0.0242014970910782\\
17	0.004511004972726\\
18	0.0140916579739738\\
19	0.00411800269594032\\
20	-0.0025107304561981\\
21	5.82993142522414e-05\\
22	-0.00266599249154305\\
23	0.00223059852998391\\
24	-0.000958949673375363\\
25	0.00412300287977123\\
26	0.00588851387713051\\
27	-0.00475807779631604\\
28	-0.0030900889605457\\
29	-0.00618113187953648\\
30	-0.00552919875116782\\
31	-0.00902710114782616\\
32	-0.00139159773497493\\
33	-0.00269203416267489\\
34	-0.00266927696992572\\
35	0.00112836184123894\\
36	-0.00135984410216987\\
37	0.002285759575373\\
38	-0.00203035236253801\\
39	-0.000479341118548253\\
40	-0.00205704972867197\\
41	-0.00527355593426055\\
42	-0.000935199167688475\\
43	-0.00472267186138812\\
44	0.00122042398158652\\
45	-0.00252892283170147\\
46	0.0155559768335156\\
47	0.0259231353595222\\
48	0.0228124764023754\\
49	0.0243767905130893\\
50	0.0264146950113769\\
51	0.0280385376187716\\
52	0.0225367961980179\\
53	0.0223488245539155\\
54	0.0242200996852557\\
55	0.0193101239161935\\
56	0.0238586199165598\\
57	0.0271148679793661\\
58	0.0187665156529276\\
59	0.0264636768492999\\
60	0.0271438596028193\\
61	0.0263359283664941\\
62	0.0203778673454423\\
63	0.0258949769702804\\
64	0.0234383729614557\\
65	0.0198289975220042\\
66	-0.018492251944114\\
67	-0.0286258951615465\\
68	-0.0228572372676541\\
69	-0.0234395681232589\\
70	-0.0243265699683642\\
71	-0.0253569119382939\\
72	-0.0273381276646179\\
73	-0.0235437760654173\\
74	-0.0261826543053536\\
75	-0.0222552346676209\\
};
\addplot [color=red,solid,line width=2.0pt,forget plot]
  table[row sep=crcr]{%
1	0.0116360868400727\\
2	0.0108881200736781\\
3	0.00677132755649959\\
4	0.013567618693558\\
5	0.0121632722563315\\
6	0.00829674619548958\\
7	0.00477291155598774\\
8	0.00591633789996359\\
9	0.0068538721671623\\
10	0.00297126109046357\\
11	0.00610952309348361\\
12	-0.00387594231309031\\
13	-0.000769946191152407\\
14	-0.00193987883084127\\
15	-0.00734111110018157\\
16	-0.00441232918595473\\
17	-0.00323075511054882\\
18	0.000646648730929116\\
19	-0.00482653293710907\\
20	-0.00377064326752416\\
21	-0.00333456294780131\\
22	-0.0050382322553292\\
23	0.000223797449140176\\
24	-0.00452969637148569\\
25	-0.00329774450842027\\
26	0.000932602626126455\\
27	0.00691381225462332\\
28	0.00140659891924348\\
29	-0.000237566730145486\\
30	-0.000202537709006753\\
31	-0.000107727943686747\\
32	0.00728174523340219\\
33	0.00270518481344677\\
34	-0.000627156502983926\\
35	0.00233819345835548\\
36	-0.000673067993105775\\
37	0.00364896295988385\\
38	-0.00482949974331226\\
39	0.00297144122155177\\
40	0.00120957307116508\\
41	-0.00037865171167866\\
42	-0.00112441443221672\\
43	-0.00136098742464\\
44	-0.00273637418304221\\
45	-0.00495041382584057\\
46	-0.00917963565480051\\
47	-0.0100780882830343\\
48	-0.00804084907321114\\
49	-0.00435815577546695\\
50	-0.00322141018688565\\
51	-0.00238170320235614\\
52	-0.00672675704848519\\
53	-0.00616592631235474\\
54	-0.00127497180021872\\
55	-0.00288667295309763\\
56	-0.00208953786236881\\
57	-0.00773791768513124\\
58	-0.00174256507643933\\
59	-0.00584119842233299\\
60	-0.00100435166982664\\
61	-0.00479154048112837\\
62	-0.00204373947124213\\
63	-0.00533908619459558\\
64	0.000935394227307906\\
65	0.00488839861707873\\
66	0.0092622275541342\\
67	0.00281044304148485\\
68	0.00624883015037512\\
69	0.00480222056293489\\
70	0.0030009129994424\\
71	0.00292988509877652\\
72	0.00345153273952719\\
73	0.00603983325218751\\
74	0.00616860765036558\\
75	0.0133157260573863\\
76	0.00719872154627513\\
77	0.00751627980947424\\
78	0.00834827578336196\\
79	0.0078228003527321\\
80	0.00255693093109884\\
81	0.00601883305228141\\
82	0.00455636273739075\\
83	0.00582676392075284\\
84	0.00566941169660226\\
85	0.00318299664955775\\
86	0.00211375947630529\\
87	0.00535111003649689\\
88	0.00637141307196718\\
89	0.00338257321139889\\
90	0.00877216556417507\\
91	0.00623689317460235\\
92	0.00650227292277275\\
93	0.00293815963165694\\
94	0.00134511587881883\\
95	-0.014548363214776\\
96	-0.0026206282705355\\
97	0.00318199822179157\\
};
\end{axis}
\end{tikzpicture}%

%% file: filterBankHist.tikz
%
%
\begin{tikzpicture}

\begin{axis}[%
width=0.4\figurewidth,
height=\figureheight,
at={(0\figurewidth,0\figureheight)},
scale only axis,
every outer x axis line/.append style={black},
every x tick label/.append style={font=\color{black}},
xmin=-0.1,
xmax=0.1,
xlabel={Bins},
xmajorgrids,
every outer y axis line/.append style={black},
every y tick label/.append style={font=\color{black}},
ymin=0,
ymax=0.25,
ylabel={Prob},
ymajorgrids,
axis background/.style={fill=white},
title={Histogram of frequencies - Scale 15},
axis x line*=bottom,
axis y line*=left
]
\addplot [color=blue,solid,forget plot]
  table[row sep=crcr]{%
-0.0592870918217354	0.0111111111111111\\
-0.0525860195786817	0\\
-0.0458849473356281	0.0111111111111111\\
-0.0391838750925745	0.0333333333333333\\
-0.0324828028495209	0.1\\
-0.0257817306064672	0.0444444444444444\\
-0.0190806583634136	0.0222222222222222\\
-0.01237958612036	0.0333333333333333\\
-0.00567851387730639	0.144444444444444\\
0.00102255836574723	0.244444444444444\\
0.00772363060880086	0.0222222222222222\\
0.0144247028518545	0.0333333333333333\\
0.0211257750949081	0.0222222222222222\\
0.0278268473379617	0.0333333333333333\\
0.0345279195810153	0.211111111111111\\
0.041228991824069	0.0333333333333333\\
};
\addplot [color=red,solid,forget plot]
  table[row sep=crcr]{%
-0.0162569394644671	0.00892857142857143\\
-0.0139581706654438	0.00892857142857143\\
-0.0116594018664205	0.00892857142857143\\
-0.00936063306739713	0.0178571428571429\\
-0.00706186426837379	0.0892857142857143\\
-0.00476309546935046	0.133928571428571\\
-0.00246432667032713	0.0714285714285714\\
-0.0001655578713038	0.107142857142857\\
0.00213321092771953	0.160714285714286\\
0.00443197972674286	0.151785714285714\\
0.0067307485257662	0.0714285714285714\\
0.00902951732478953	0.0535714285714286\\
0.0113282861238129	0.0267857142857143\\
0.0136270549228362	0.0535714285714286\\
0.0159258237218595	0.0267857142857143\\
0.0182245925208829	0.00892857142857143\\
};
\addplot [color=blue,solid,line width=2.0pt,forget plot]
  table[row sep=crcr]{%
-0.0592870918217354	0.0111111111111111\\
-0.0525860195786817	0\\
-0.0458849473356281	0.0111111111111111\\
-0.0391838750925745	0.0333333333333333\\
-0.0324828028495209	0.1\\
-0.0257817306064672	0.0444444444444444\\
-0.0190806583634136	0.0222222222222222\\
-0.01237958612036	0.0333333333333333\\
-0.00567851387730639	0.144444444444444\\
0.00102255836574723	0.244444444444444\\
0.00772363060880086	0.0222222222222222\\
0.0144247028518545	0.0333333333333333\\
0.0211257750949081	0.0222222222222222\\
0.0278268473379617	0.0333333333333333\\
0.0345279195810153	0.211111111111111\\
0.041228991824069	0.0333333333333333\\
};
\addplot [color=red,solid,line width=2.0pt,forget plot]
  table[row sep=crcr]{%
-0.0162569394644671	0.00892857142857143\\
-0.0139581706654438	0.00892857142857143\\
-0.0116594018664205	0.00892857142857143\\
-0.00936063306739713	0.0178571428571429\\
-0.00706186426837379	0.0892857142857143\\
-0.00476309546935046	0.133928571428571\\
-0.00246432667032713	0.0714285714285714\\
-0.0001655578713038	0.107142857142857\\
0.00213321092771953	0.160714285714286\\
0.00443197972674286	0.151785714285714\\
0.0067307485257662	0.0714285714285714\\
0.00902951732478953	0.0535714285714286\\
0.0113282861238129	0.0267857142857143\\
0.0136270549228362	0.0535714285714286\\
0.0159258237218595	0.0267857142857143\\
0.0182245925208829	0.00892857142857143\\
};
\end{axis}

\begin{axis}[%
width=0.4\figurewidth,
height=\figureheight,
at={(0.54\figurewidth,0\figureheight)},
scale only axis,
every outer x axis line/.append style={black},
every x tick label/.append style={font=\color{black}},
xmin=-0.04,
xmax=0.04,
xlabel={Bins},
xmajorgrids,
every outer y axis line/.append style={black},
every y tick label/.append style={font=\color{black}},
ymin=0,
ymax=0.18,
ylabel={Prob},
ymajorgrids,
axis background/.style={fill=white},
title={Histogram of frequencies - Scale 20},
axis x line*=bottom,
axis y line*=left
]
\addplot [color=blue,solid,forget plot]
  table[row sep=crcr]{%
-0.0286258951615465	0.0266666666666667\\
-0.0242254187253557	0.0933333333333333\\
-0.0198249422891649	0.0133333333333333\\
-0.0154244658529741	0\\
-0.0110239894167833	0.0133333333333333\\
-0.00662351298059247	0.0666666666666667\\
-0.00222303654440166	0.173333333333333\\
0.00217743989178915	0.133333333333333\\
0.00657791632797996	0.04\\
0.0109783927641708	0.0266666666666667\\
0.0153788692003616	0.0266666666666667\\
0.0197793456365524	0.0533333333333333\\
0.0241798220727432	0.173333333333333\\
0.028580298508934	0.133333333333333\\
0.0329807749451248	0.0133333333333333\\
0.0373812513813156	0.0133333333333333\\
};
\addplot [color=red,solid,forget plot]
  table[row sep=crcr]{%
-0.014548363214776	0.0103092783505155\\
-0.012673964420887	0\\
-0.0107995656269981	0.0103092783505155\\
-0.00892516683310916	0.0206185567010309\\
-0.00705076803922023	0.0412371134020619\\
-0.0051763692453313	0.103092783505155\\
-0.00330197045144236	0.103092783505155\\
-0.00142757165755343	0.11340206185567\\
0.000446827136335501	0.103092783505155\\
0.00232122593022443	0.134020618556701\\
0.00419562472411337	0.0721649484536082\\
0.0060700235180023	0.154639175257732\\
0.00794442231189123	0.0721649484536082\\
0.00981882110578016	0.0103092783505155\\
0.0116932198996691	0.0309278350515464\\
0.013567618693558	0.0206185567010309\\
};
\addplot [color=blue,solid,line width=2.0pt,forget plot]
  table[row sep=crcr]{%
-0.0286258951615465	0.0266666666666667\\
-0.0242254187253557	0.0933333333333333\\
-0.0198249422891649	0.0133333333333333\\
-0.0154244658529741	0\\
-0.0110239894167833	0.0133333333333333\\
-0.00662351298059247	0.0666666666666667\\
-0.00222303654440166	0.173333333333333\\
0.00217743989178915	0.133333333333333\\
0.00657791632797996	0.04\\
0.0109783927641708	0.0266666666666667\\
0.0153788692003616	0.0266666666666667\\
0.0197793456365524	0.0533333333333333\\
0.0241798220727432	0.173333333333333\\
0.028580298508934	0.133333333333333\\
0.0329807749451248	0.0133333333333333\\
0.0373812513813156	0.0133333333333333\\
};
\addplot [color=red,solid,line width=2.0pt,forget plot]
  table[row sep=crcr]{%
-0.014548363214776	0.0103092783505155\\
-0.012673964420887	0\\
-0.0107995656269981	0.0103092783505155\\
-0.00892516683310916	0.0206185567010309\\
-0.00705076803922023	0.0412371134020619\\
-0.0051763692453313	0.103092783505155\\
-0.00330197045144236	0.103092783505155\\
-0.00142757165755343	0.11340206185567\\
0.000446827136335501	0.103092783505155\\
0.00232122593022443	0.134020618556701\\
0.00419562472411337	0.0721649484536082\\
0.0060700235180023	0.154639175257732\\
0.00794442231189123	0.0721649484536082\\
0.00981882110578016	0.0103092783505155\\
0.0116932198996691	0.0309278350515464\\
0.013567618693558	0.0206185567010309\\
};
\end{axis}
\end{tikzpicture}%